# Remaining Useful Life Prediction for Batteries Utilizing an Explainable AI Approach with a Predictive Application for Decision-Making

Biplov Paneru[a*], Bipul Thapa[b], Durga Prasad Mainali[cd*], Bishwash Paneru[e], Krishna Bikram Shah[g]

[a]Department of Electronics and Communication Engineering, Nepal Engineering College, Pokhara University, Bhaktapur Nepal

[b]Department of Computer and Information Sciences, University of Delaware, Newark, USA

[c]Department of Mechanical, Maritime and Materials Engineering, Delft University of Technology, Delft, Netherlands

[d]Department of Technology, Innovation and Society, The Hague University of Applied Science, The Hague, Netherlands

[e]Department of Applied Sciences and Chemical Engineering, Institute of Engineering, Pulchowk Campus, Tribhuvan University, Lalitpur, Nepal

[g]Department of Computer Science and Engineering, Nepal Engineering College, Pokhara University, Bhaktapur Nepal

*Corresponding author: Biplov Paneru, Durga Prasad Mainali

**corresponding author 1 email:** biplov001@gmail.com

**corresponding author 2 email**: D.P.Mainali@tudelft.nl

**Abstract:**

Accurately estimating the Remaining Useful Life (RUL) of a battery is essential for determining its lifespan and recharge requirements. In this work, we develop machine learning-based models to predict and classify battery RUL. We introduce a two-level ensemble learning (TLE) framework and a CNN+MLP hybrid model for RUL prediction, comparing their performance against traditional, deep, and hybrid machine learning models. Our analysis evaluates various models for both prediction and classification while incorporating interpretability through SHAP. The proposed TLE model consistently outperforms baseline models in RMSE, MAE, and $R^2$, demonstrating its superior predictive capabilities. Additionally, the XGBoost classifier achieves an impressive 99% classification accuracy, validated through cross-validation techniques. The models effectively predict relay-based charging triggers, enabling automated and energy-efficient charging processes. This automation reduces energy consumption and enhances battery performance by optimizing charging cycles. SHAP interpretability analysis highlights the cycle index and charging parameters as the most critical factors influencing RUL. To improve accessibility, we developed a Tkinter-based GUI that allows users to input new data and predict RUL in real time. This practical solution supports sustainable battery management by enabling data-driven decisions about battery usage and maintenance, contributing to energy-efficient and innovative battery life prediction.

## 1. Introduction

The rapid adoption of battery-powered systems in modern technology, from electric vehicles to renewable energy storage, has highlighted the critical need for accurate and efficient battery management. Prolonged use and improper maintenance can severely impact battery performance, leading to inefficiencies, unexpected failures, and increased costs [31]. A key challenge in battery management lies in accurately estimating the Remaining Useful Life (RUL), which provides an essential measure of how long a battery can function reliably. Factors such as temperature fluctuations, chemical degradation, and charge-discharge cycles contribute to the steady decline in a battery's energy storage capacity [32]. Without precise RUL predictions, systems are vulnerable to unplanned breakdowns, safety risks, and inefficient maintenance practices. These challenges hinder the development of sustainable energy systems and drive the demand for more reliable and interpretable solutions. The motivation for this work stems from the increasing importance of sustainable energy solutions and the potential of machine learning to address existing limitations in battery management. Accurate RUL prediction enables proactive maintenance, optimizes replacement schedules, and minimizes unexpected failures. Furthermore, interpretability in machine learning models is crucial for understanding key factors influencing battery performance, allowing users to make data-driven decisions. Additionally, integrating automation into charging systems offers a path to reduce energy waste and enhance battery performance.

This research proposes a two-level ensemble learning (TLE) framework and a CNN+MLP hybrid model for RUL prediction to address these challenges. The two-level ensemble learning framework combines two groups of models, where each group stacks multiple algorithms to optimize predictions. The final prediction is achieved by aggregating the outputs of these groups through a bagging approach, ensuring robust and accurate RUL estimation. For the classification, we use the XGBoost and MLP models. These models are augmented by interpretability analysis using LIME to identify critical features influencing RUL estimation. To ensure accessibility, a Tkinter-based GUI is developed, enabling users to input new data and predict RUL in real time. The proposed approach significantly impacts the sustainability and efficiency of battery management. By improving the accuracy and interpretability of RUL predictions, the solution empowers users to optimize maintenance and reduce energy consumption. Automation of charging processes further enhances battery longevity and minimizes waste, contributing to a more sustainable energy ecosystem. This work advances the fields of machine learning, automation, and energy management, paving the way for innovative and sustainable solutions in battery-powered systems.

## Literature Review

This section delves into the advancements in battery state prediction techniques, emphasizing the role of AI and ML in enhancing the accuracy and efficiency of SOC and SOH estimations. The integration of these advanced methods addresses critical challenges such as dynamic load prediction, real-time performance assessment, and long-term battery health monitoring.

Ng et al. point [1] out the main difficulties, particularly in executing in situ computations, high-throughput data gathering, and accurate modeling over extended periods of time. All things considered, the work offers insights into explainable, real-time machine learning for future battery production, administration, and optimization. Shan et al. review the most recent findings on widely used ML techniques for predicting SOC and SOH, giving a thorough overview of both BMSs and ML. It also emphasizes the difficulties involved. This research highlights the prevalence of a support vector machine (SVM), fuzzy logic (FL), k-nearest neighbors (KNN) algorithm, genetic algorithm (GA), and transfer learning in SOC and SOH estimates, in addition to more conventional models such equivalent circuit models (ECMs) and electrochemical battery models. Narayanan et al. performed the analysis of the suggested techniques using real-time Lithium Ion battery data at various temperature profiles. Their study takes into account the R-Square () and Root Mean Square Error (RMSE) indices to verify the effectiveness of the suggested approach. The study's findings indicate that the neural network-based prediction method—which has both high and low RMSE indices—is the best one. The advantages of the suggested strategy include its great accuracy and simplicity. Applications for electric vehicles can make use of this anticipated battery model.

Jayakumar et al. [4] used the ensemble random forest model to minimize data degradation for RL prediction. The model makes it possible to gather data and use random forest and ensemble random forest for preprocessing and classification. They achieved a higher accuracy in preduction using the ensemble random forest model. Oyucu et al. proposed a methodology for comparative analysis that focuses on deep learning and classical approaches along with the discussion of the enhancements to the LSTM and BiLSTM models By forecasting LIB performance, the study hopes to further technological progress in the electric vehicle sector. Qaisar et al. proposed a feature extraction methodology from battery charge/discharge curves using statistical analysis and shape context, enabling effective State of Charge (SoC) prediction. By implementing and benchmarking decision trees, random forests, and linear regression in MATLAB, the random forest regressor demonstrates superior performance with a correlation coefficient of 0.9988, showcasing its robustness.

Dineva et al. [7] evaluated the performance of advanced machine learning techniques for SoC prediction under dynamic load conditions using regression models. A dynamic charge/discharge dataset was generated through a unique multisine signal-based testing approach. Their findings highlight that a key advantage of advanced ML models lies in their ability to uncover critical correlations between relevant variables. These state-of-the-art techniques outperform traditional ML methods by effectively capturing battery cell dynamics and leveraging historical data, making them highly suitable for accurate SoC forecasting. Su et al. demonstrated an innovative methodology combining feature extraction with multiple linear regression to predict a complete battery charge curve using only a fraction of the input data. This approach achieves a prediction error of less than 2% when using just 10% of the charge curve as input, and its effectiveness is further validated on LiCoO2-based batteries, maintaining the same level of accuracy with only 5% of input data. This highlights the technique's generalizability and potential for real-world applications, offering a quick and efficient means for onboard battery health monitoring and estimation. Such contributions underscore the growing importance of advanced data-driven models in battery health diagnostics and prediction tasks, providing valuable context for the current study's focus on predictive modeling.

Characterizing the uncertainty in a model's predictions is crucial for making well-informed decisions on field control tactics or lab battery design. Thelen et al. analyzed the state-of-the-art probabilistic machine learning models for health diagnostics and prognostics, after giving an overview of lithium-ion battery degradation. There is a thorough discussion of the different approaches, their benefits, and drawbacks, with a major focus on probabilistic machine learning and uncertainty quantification. Finally, prospects for research and development as well as future trends are explored. Haripriya et al. [10] used sensor data from the LPC2148 ARM board, including voltage, current, and temperature parameters for various algorithms such as LSTM, Decision Tree (DT), K-Nearest Neighbors (KNN), Naïve Bayes (NB), and Support Vector Machine (SVM). Among these, the Naïve Bayes algorithm demonstrated superior performance on real-time data, achieving the highest F1-score, accuracy, precision, and recall. With an accuracy of 88%, Naïve Bayes proved effective in estimating the Remaining Battery Capacity, contributing to the prediction of lithium-ion battery aging. This study highlights the potential of lightweight algorithms for real-time battery health monitoring. Kawahara et al. proposed an evaluation metric based on the average voltage of the test data plus three standard deviations, validated using the root mean squared error (RMSE) of voltage at various lower OCV limits. Among eight machine learning models evaluated, the multilayer perceptron (MLP) demonstrated the best extrapolation accuracy, achieving 92.7 mV. Using published experimental data, the MLP further exhibited an accuracy of 102.4 mV, confirming its superior performance for battery voltage extrapolation tasks.

Sapra et al. [12] conducted tests on lithium polymer battery cells to evaluate performance metrics such as voltage, current, and battery capacity. They developed physics-based and machine-learning models to forecast the State of Charge (SoC), using high C-rate measurements (1C to 4C) for testing, calibration, and training. Their findings showed that the Pseudo-2D electrochemical model estimated SoC with a root mean squared error (RMSE) of approximately 2% across various C-rates. However, the Feed Forward Neural Network approach, coupled with Butterworth and Hampel filters, achieved even greater accuracy, with RMSE values of less than or around 1%, demonstrating its superior performance for precise SoC prediction. Paneru et al. [28] used machine learning models, including gradient boosting, random forests, decision trees, and linear regression, to predict charging cycles in electric vehicle (EV) battery systems. The Gradient Boosting model outperformed others with an R-squared score of 0.87, followed by Random Forest at 0.83. These findings emphasize the critical role of model selection in enhancing prediction accuracy. Additionally, the study resulted in the creation of an EV Battery Charging Cycle Predictor App, demonstrating the potential of advanced machine learning techniques to improve EV battery efficiency, maintenance scheduling, and energy decision-making in electric mobility technologies. Chen J. (2013) [34] charts the development of Li-ion battery materials. Because of their excellent reversibility and safety characteristics, carbon-based materials were developed as anode materials in the 1990s. Although carbon-based anodes are inexpensive and have a long cycle life, their limited capacity makes the investigation of substitute materials, including lithium-tin alloys, necessary.

Extensive research efforts have focused on developing sustainable power management systems [1-13] and alternative fuel solutions [14-25], particularly for vehicle and industrial applications. However, a critical gap persists in accurately predicting battery lifespan and automating energy processes such as battery charging. With the integration of explainable AI technology with a GUI application, this work presents an integrated system for predicting and automating battery charging with extensive use of different Neural Network models like CNN to study and research more on the energy management-related field.

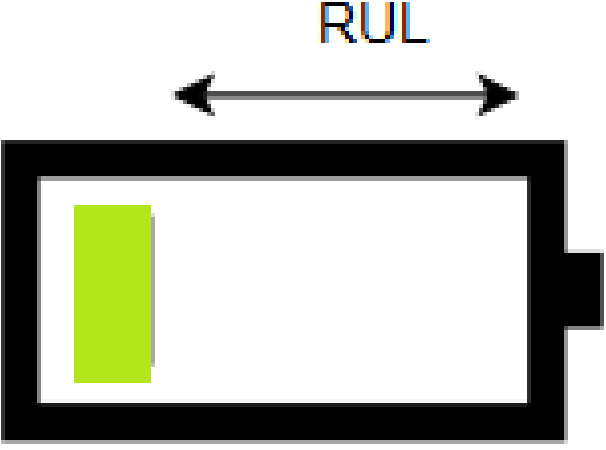

**Fig. 1.** Battery RUL

## 2. Methodology

The various features are analyzed to implement machine learning-based models that can predict the RUL of the vehicle, with the aid of various parameters like charging cycles, discharge time, maximum voltage, etc. the RUL and finally, SHAP explainable AI is utilized for this purpose of the most affecting feature investigation.

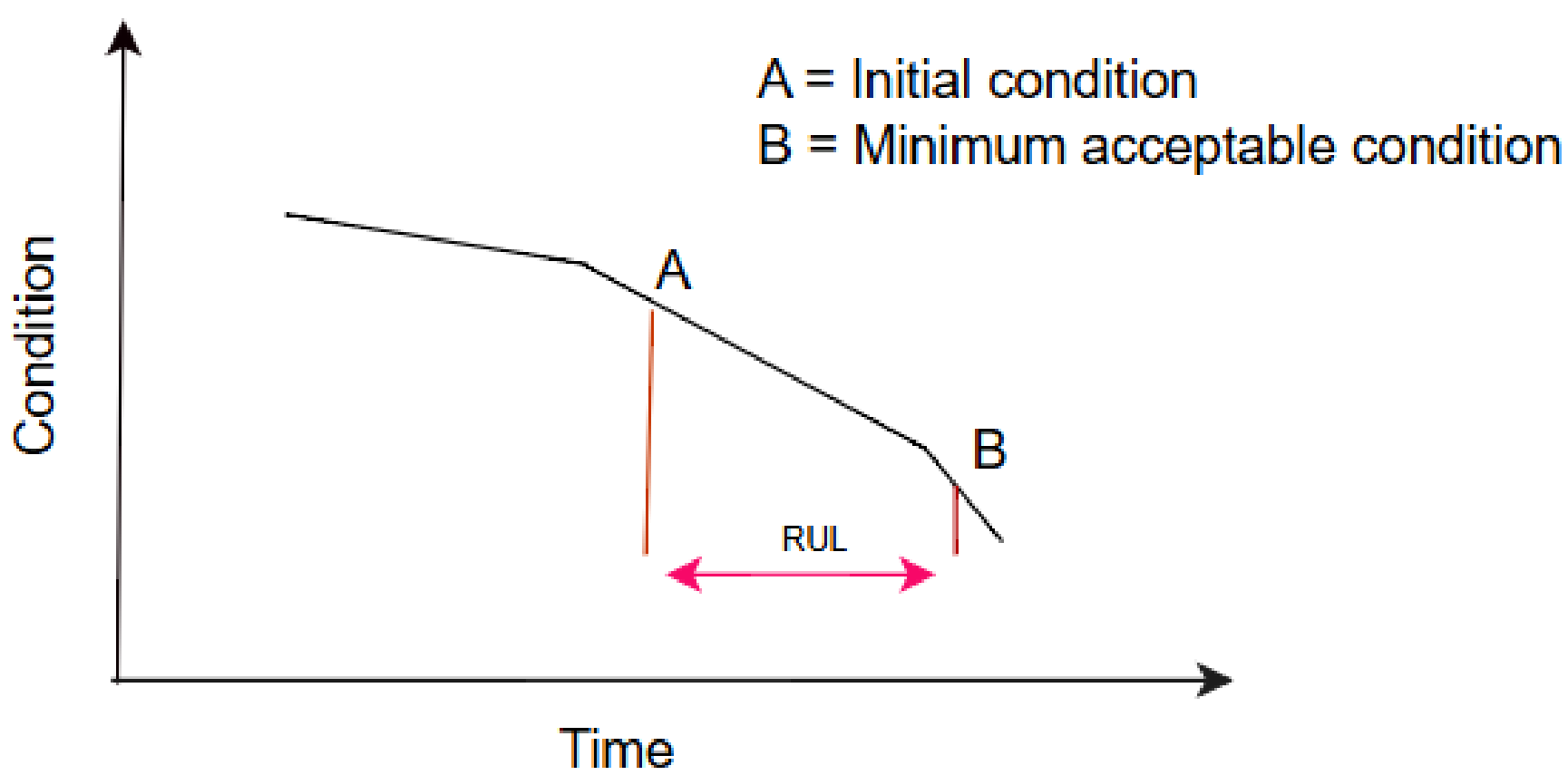

**Fig 2.** Graphical representations of charging conditions

A battery's estimated RUL is the number of cycles or amount of time it may be used efficiently before reaching the end of its operational life. In battery-powered systems, such as electric vehicles and renewable energy storage, the RUL is a crucial metric for anticipating when a battery will no longer meet the required performance standards. Figure 1. depicts how a machine's performance gradually deteriorates over time. The RUL is the amount of time between point A, which represents the machine's current state, and point B, which represents the moment at which the machine can no longer operate as intended. Predicting the RUL in advance allows for the scheduling of maintenance or replacement, preventing unplanned malfunctions and monetary losses. Precise projections are vital since calculating RUL accurately is necessary for efficient operations and well-informed decision-making.

### *2.1 Dataset*

Fourteen NMC-LCO 18650 batteries with a nominal capacity of 2.8 Ah were tested by the Hawaii Natural Energy Institute. The batteries were cycled more than a thousand times at 25°C using a CC-CV charge rate of C/2 rate and a discharge rate of 1.5C. The dataset obtained from Kaggle [26] contained the following features:

i. Cycle Index: number of cycles
ii. F1: Discharge Time (s)
iii. F2: Time at 4.15V (s)
iv. F3: Time Constant Current (s)
v. F4: Decrement 3.6-3.4V (s)
vi. F5: Max. Voltage Discharge (V)
vii. F6: Min. Voltage Charge (V)
viii. F7: Charging Time (s)
ix. Total time (s)
x. RUL: target variable

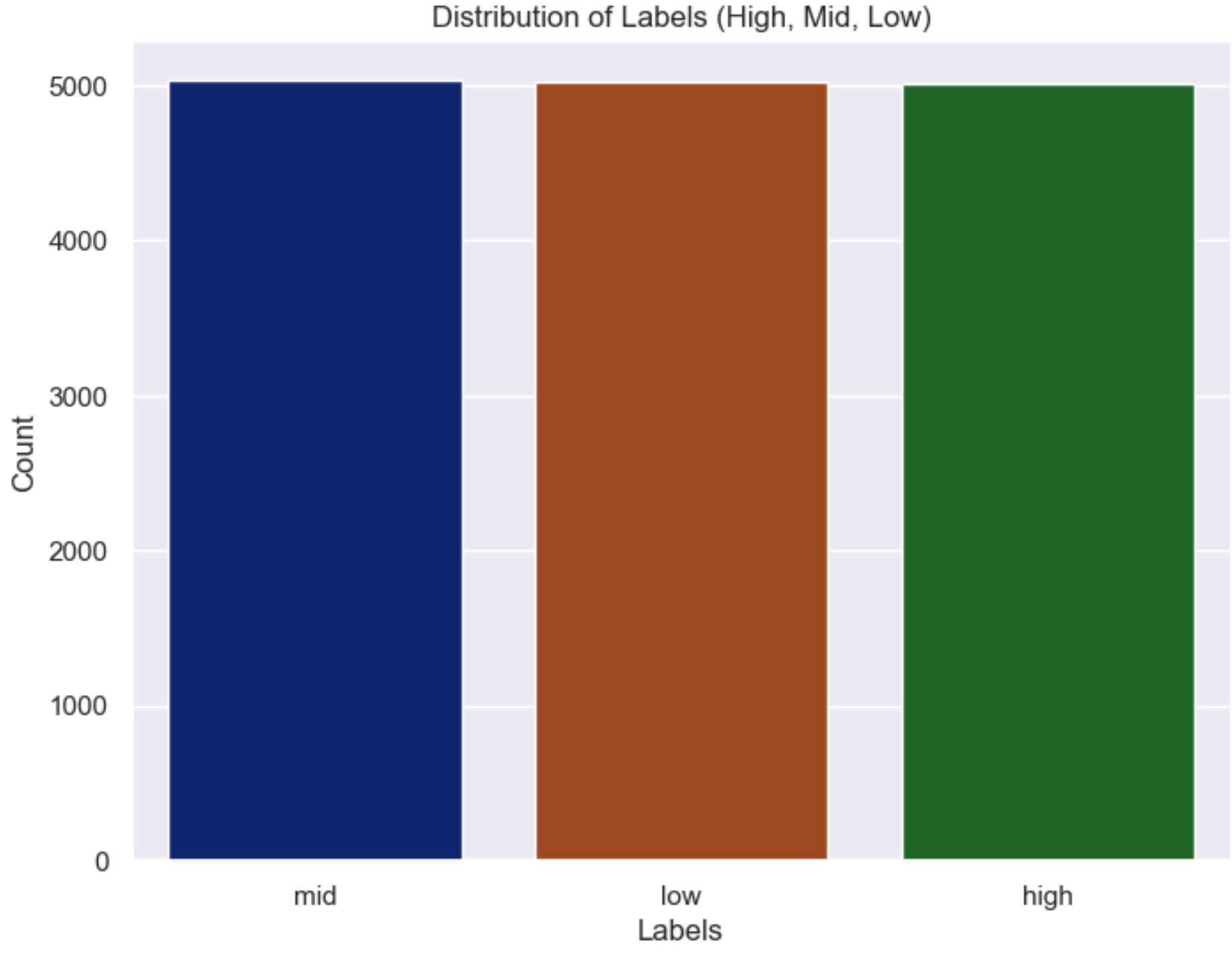

**Fig 3.** Dataset distribution

### *2.2 Dataset Preprocessing*

The dataset was utilized with the last column RUL values were distinguished into 3 classes.

Table 1. Dataset distribution phenomenon

| **RUL level** | **Class** |
|---|---|
| <= 369 | Low |
| >369 <= 748 | Mid |
| >748 | High |

The evaluation of the dataset revealed no presence of outliers, indicating that the data was sufficiently clean for model development. No significant outliers and unwanted data were present, so the dataset was very suitable to be utilized for the predicting and modeling. The dataset classes were created for the classification approach and were labeled as per Table 1, the concept of class balance was too nearly maintained with this equalized distribution of the dataset.

### **2.3 Features Extraction**

In the last column, the RUL label column was updated with their division into classes and categories, which helped in classifying the RUL values. The dataset was developed with features that highlight the voltage and current behavior across each cycle using that source dataset. The RUL of the batteries can be estimated using those features. The 14 batteries' synopsis is included in the dataset.

Additionally, we introduce three additional features to capture critical battery characteristics and operational patterns. These features include Total Time, Charge-to-Discharge Ratio, and Voltage Range. The Total Time feature is computed as the sum of discharge and charging times, representing the overall operational duration of the battery. The Charge-to-Discharge Ratio is introduced as the ratio of charging time to discharge time, reflecting the efficiency and balance of the battery's charge cycle. Finally, the Voltage Range is calculated as the difference between the maximum discharge voltage and minimum charging voltage, capturing the range of voltage fluctuations during operation.

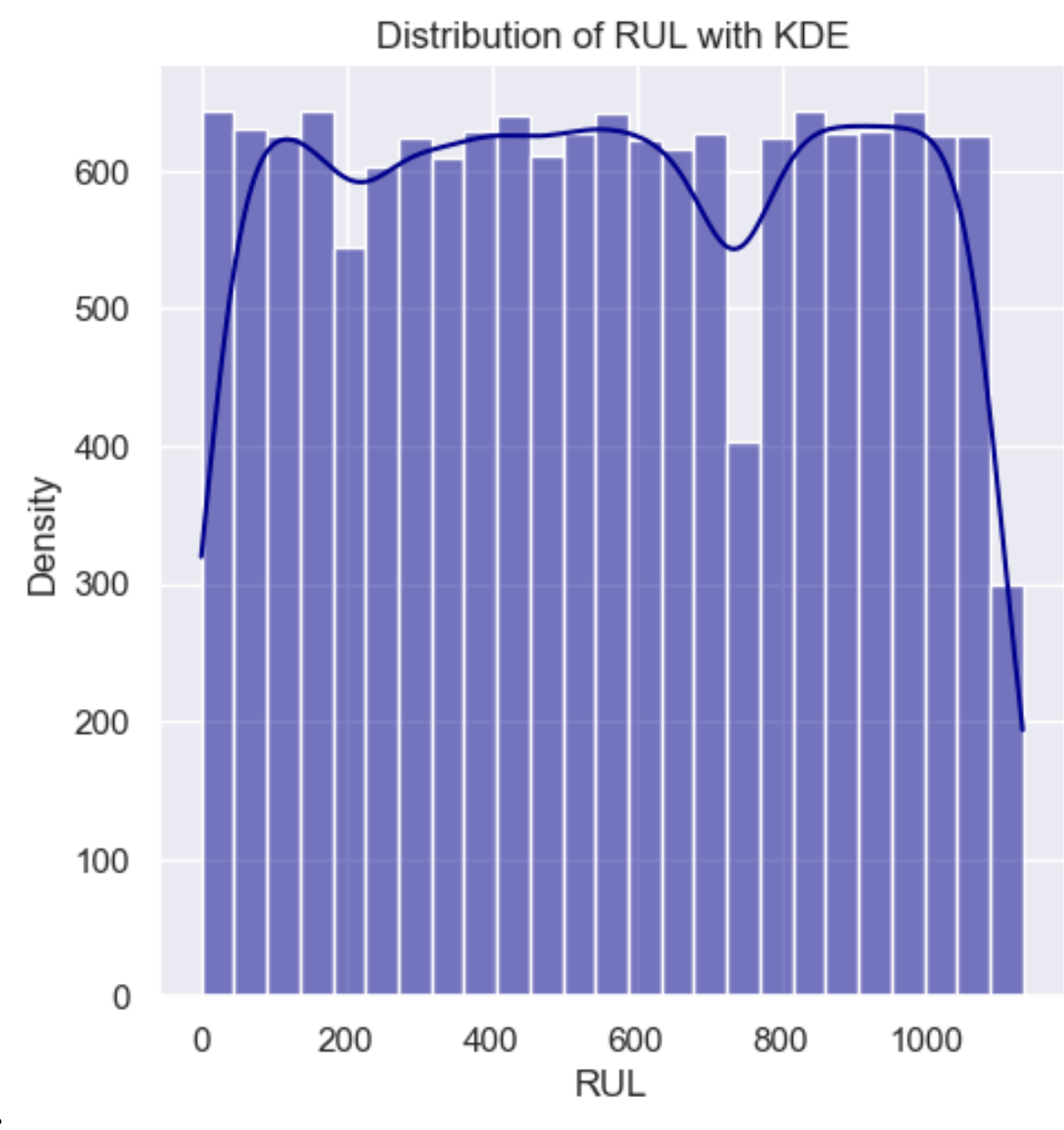

**Fig 4.** RUL distribution plot with KDE

The plot in Figure 4 presents the distribution of the batteries' RULs, illustrated through a histogram with an overlaid Kernel Density Estimate (KDE) for better visualization. The frequency of RUL values over various intervals seems to be consistently distributed between 0 and 1000, indicating a broad range of battery life expectancy within the dataset. A continuous representation of the underlying distribution is provided by the KDE line, which offers a smooth estimate of the probability density function of the RUL [33]. The density peaks between 600 and 700, then slightly declines to about 500, and then rises again close to the maximum RUL. There is an upper limit to the projected battery longevity in this dataset, as indicated by the drop-off at the end, which shows that fewer batteries have an RUL near the maximum value. The distribution's general shape is somewhat flat, suggesting that RUL values are distributed rather evenly, with a sizable percentage of batteries falling inside the whole RUL range.

### *2.4 Splitting the Dataset*

The dataset is split into training and testing sets in an 80:20 ratio. Additionally, for deep learning models, 20% of the training set is further allocated as a validation set.

### *2.5 Proposed Workflow*

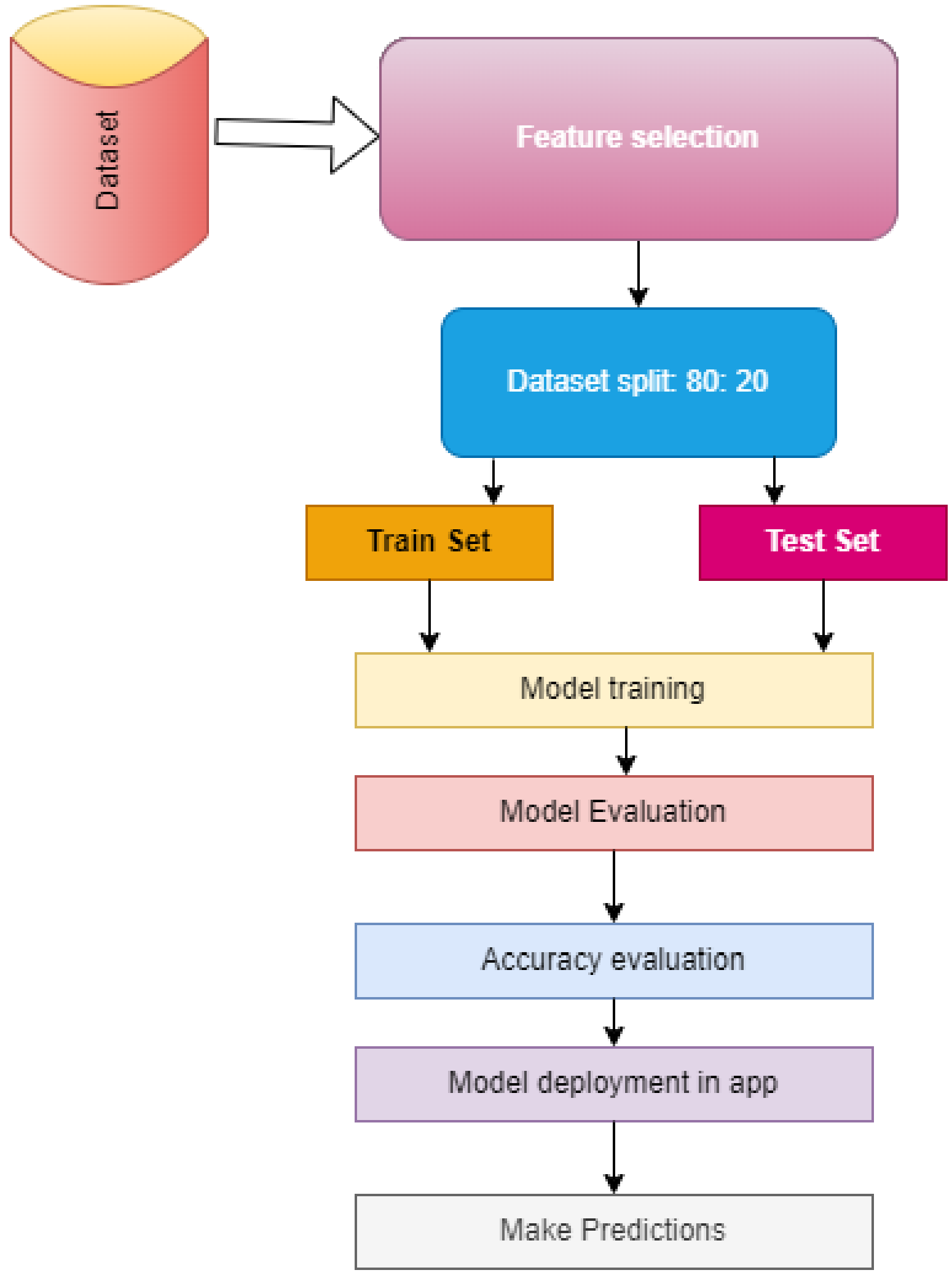

**Fig 5.** Proposed workflow

The proposed workflow of the given work is depicted in Figure 5. The dataset is collected, outliers are checked, then the labels are created based on RUL values for the classification approach, and the model is trained in various algorithms that are finally deployed on the Tkinter GUI application to make predictions. Different machine learning models are developed, fine-tuned, and evaluated for best performance.

### *2.6 Evaluation metrics*

i. Accuracy: The percentage of right guesses among all the forecasts made is the measure of accuracy. Although it is a widely used statistic for classification algorithms, an imbalanced dataset leads to deceptive results.

Accuracy = $\frac{TP+TN}{(FP+FN+TP+TN)}$…………..(1)

Where TP = True Positives, TN = True Negatives, FP = False Positive, FN = False Negative

ii. Precision:

Precision, which is the ratio of true positive forecasts to all positive predictions, is sometimes referred to as positive predictive value. It shows how well the model predicts the good outcomes.

Precision=$\frac{TP}{(FP+TP)}$ ……….(2)

iii. Recall:

Recall quantifies the percentage of real positive cases that the model properly recognized; it is sometimes referred to as Sensitivity or True Positive Rate. It shows how well the model can capture good examples.

Recall = $\frac{TP}{(FN+TP)}$……………….(3)

iv. F1-Score

Precision and recall are harmonic means, and the F1 score strikes a balance between both. When the dataset is unbalanced and both false positives and false negatives must be taken into account, it is helpful.

F1 Score=2 $\times \frac{Precision*Recall}{Precision+Recall}$………..(4)

v. **$R^2$ Score:**

A statistical metric known as the coefficient of determination, or $R^2$ score, shows how well a regression model fits the data. It shows the percentage of the dependent variable's (goal) volatility that can be predicted based on the independent variables (features). A model that accurately predicts the target has an $R^2$ score of 1, whereas a model that does not capture any variability in the data has a value of 0. Essentially, a regression model's goodness of fit is evaluated using the R2 score, where higher values correspond to greater prediction performance.

a. $$R^2 = \frac{\sum_{a=1}^{n}(x_a - x_m)^2 - \sum_{a=1}^{n}(x_a - y_a)^2}{\sum_{a=1}^{n}(x_a - x_m)^2} \quad \text{……………………..} \quad (5)$$

vi. **RMSE:**

The average size of errors between the predicted and actual values is measured by the Root Mean Squared Error (RMSE), a metric used to assess a regression model's accuracy. The square root of the average of the squared discrepancies between the actual values and the projections is what it is. Better model performance is indicated by lower RMSE values, which show how concentrated the data is around the line of best fit.

$$RMSE = \sqrt{\frac{1}{n}\sum_{a=1}^{n}(x_a - y_a)} \text{ ........................ (6)}$$

Where, $x_a$, $y_a$, and $x_m$ are observed, measured, and average values for each n observations.

**vii. MAE:**

The Mean Absolute Error (MAE) measures the average magnitude of errors between predicted and actual values in a regression model. It is calculated as the average of the absolute differences between predicted and actual values. Lower MAE values indicate better model performance, reflecting how close predictions are to the actual values.

$$MAE = \frac{1}{n}\sum_{a=1}^{n} |x_a - y_a| \text{........................ (6)}$$

### 2.7 Cross-validation

Nested cross-validation is a robust approach for model evaluation and hyperparameter tuning, consisting of an outer loop for performance assessment on unseen data and an inner loop for optimizing hyperparameters through cross-validation. This method minimizes overfitting and enhances generalization. We used 5-fold stratified cross-validation for regression and 10-fold nested cross-validation for RUL classification.

### 2.8 Proposed Model

Various traditional, ensemble, and deep learning models are selected to predict various classes as well as RUL values and the key motivation is to surpass the performance of state-of-the-art models.

**i. MLP+CNN model:**

This hybrid model can handle both structured and unstructured data types since it includes a Convolutional Neural Network (CNN) with a Multi-Layer Perceptron (MLP). While the CNN layers are very good at tasks involving spatial patterns, like image data, the MLP layers are better at capturing intricate relationships in the data. In situations where it's necessary to understand both local and global data patterns, this architecture can be helpful. The model is utilized here for predicting RUL utilizing a regression algorithm.

### ii. Two-Level Ensemble (TLE):

The TLE model is specifically designed to handle complex structured data by combining multiple machine-learning techniques through stacking and bagging as shown in Fig 7. The first level consists of two groups: simpler models like Decision Trees, Extra Tree Regressor, and Random Forests, and advanced boosting models like XGBoost, LightGBM, and CatBoost. These models excel in capturing both simple and complex patterns in the data. The second level aggregates predictions from these groups using a meta-model and combines their outputs through bagging, which enhances the robustness and generalization of the predictions. By leveraging the strengths of different algorithms and ensemble techniques, this model is highly effective for tasks that require understanding intricate relationships and interactions within the data. The model is utilized here for predicting RUL using advanced regression algorithms, ensuring high accuracy and reliability in performance.

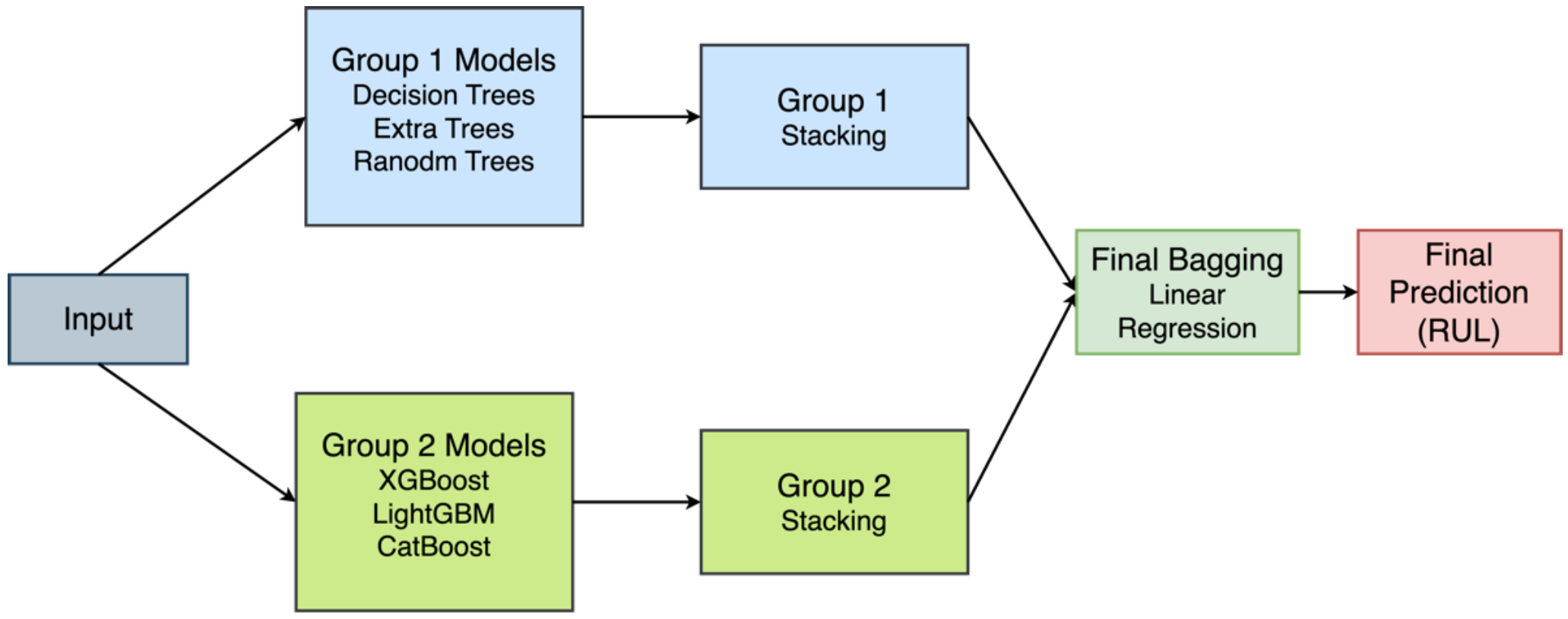

**Fig 7.** TLE architecture

### 2.9 Baseline

For the baseline comparison, we use the XGBoost, CatBoost, BiGRU, and CNN-BiLSTM [34] models. The XGBoost and CatBoost are gradient-boosting algorithms that are efficient and perform well for numerical data, leveraging decision tree ensembles to capture feature interactions and deliver strong predictive performance. BiGRU and CNN-BiLSTM are deep neural network models that use sequential layers and recurrent structures to model temporal dependencies and complex patterns in time-series data. BiGRU captures information from both past and future time steps, making it highly effective for sequential tasks, while CNN-BiLSTM combines convolutional layers for feature extraction and bidirectional LSTM layers for learning long-term dependencies. The overall model usage in our work is depicted in Table 2.

Table 2: Implemented Models

| **Type** | **Models** |
|---|---|
| Regression | XGBoost |
| | CatBoost |
| | BiGRU |
| | CNN-BiLSTM |
| | MLP + CNN |
| | TLE |
| Classification | XGBoost |
| | MLP |

## 3 Results & Discussions

The various models were developed with the main goal of predicting and classifying RUL, performing interpretability analysis with SHAP, and finally, utilizing a GUI to predict the RUL for the battery.

### 3.1 Results of the Developed Model

***XGboost:***

The plot in Fig. 8 compares the predicted RUL values generated by the XGBoost regression mod el with the actual RUL values. The diagonal red line represents perfect predictions, and the cluste ring of most data points (blue dots) along this line demonstrates the model's high accuracy. This alignment is particularly evident in the mid-to-higher RUL ranges, indicating minimal errors in t hese regions. The model's performance is further validated by quantitative metrics, including an MAE of 4.4955, an RMSE of 8.0441, and an $R^2$ score of 0.9994, showing a near-perfect correlati on between predictions and actual values. However, a slight dispersion of points at lower RUL v alues suggests minor prediction errors, potentially caused by noise or data imbalance in these ran ges.

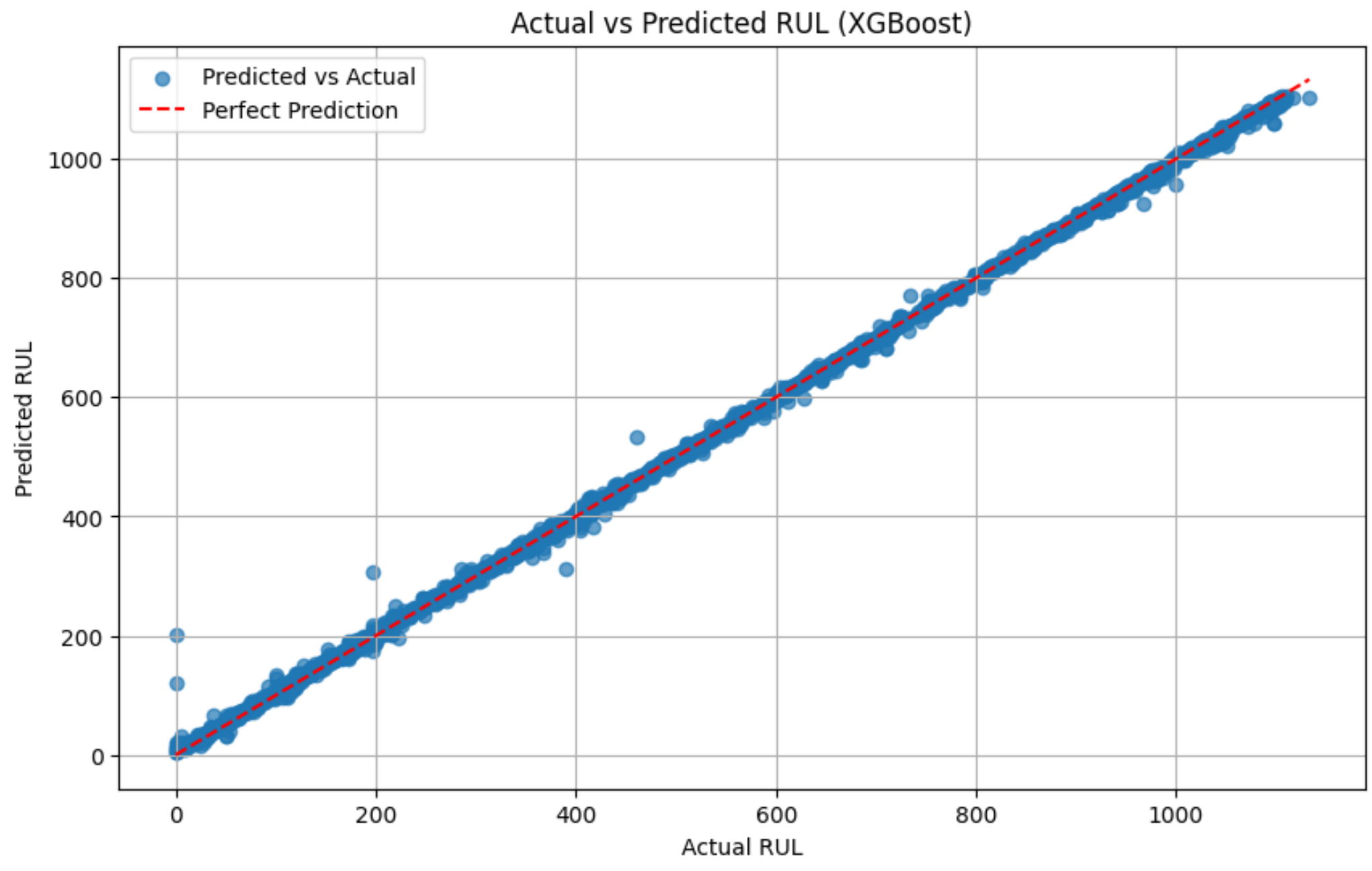

Figure 8: Actual vs predicted vs XGboost regressor model

***Catboost:***

The model catboost got an excellent $R^2$ Score: 0.9997 and similarly, a RMSE of: 5.1060

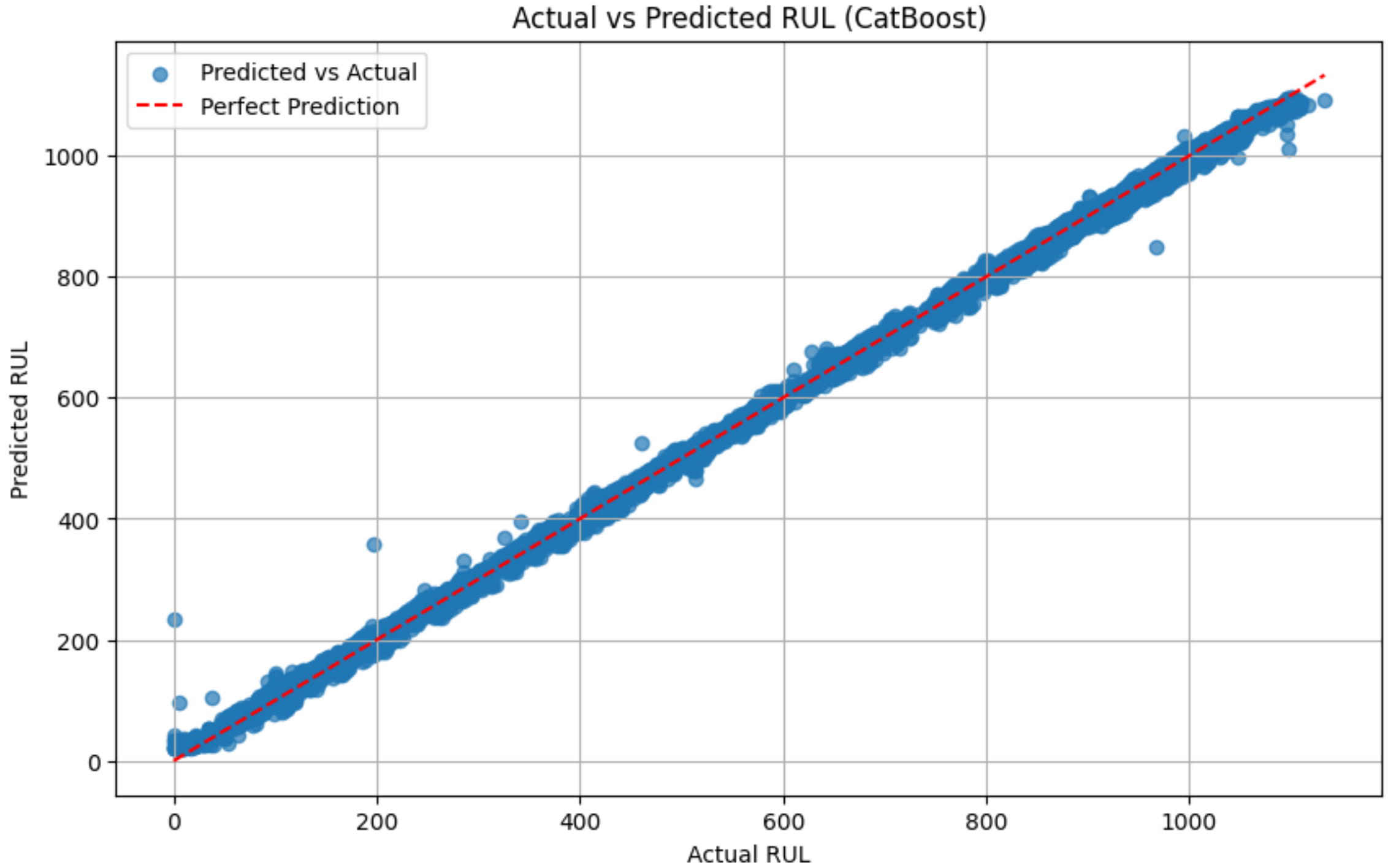

Figure 9: Actual vs predicted plot for CatBoost model

The plot in Fig 9 displays a comparison between the predicted RUL values produced by the CatBoost regressor model and the actual RUL values. Most of the data points (blue dots) align closely with this line, indicating a strong correlation and accurate predictions, particularly in the mid-to-higher RUL ranges. The model's performance is further evaluated using quantitative metrics, yielding an MAE of 10.6077, an RMSE of 14.3522, and an $R^2$ score of 0.9980.

***MLP+CNN***

The CNN and MLP hybridized model achieved a final RMSE of 12.630 and an $R^2$ score of 0.9985, demonstrating its strong predictive performance. As shown in Fig 10 (left), the actual vs. predicted RUL graph illustrates that most data points lie close to the diagonal red line representing per fect predictions, indicating the model’s high accuracy in estimating RUL. The alignment of point s along the diagonal reflects the model's ability to generalize well for the majority of samples.

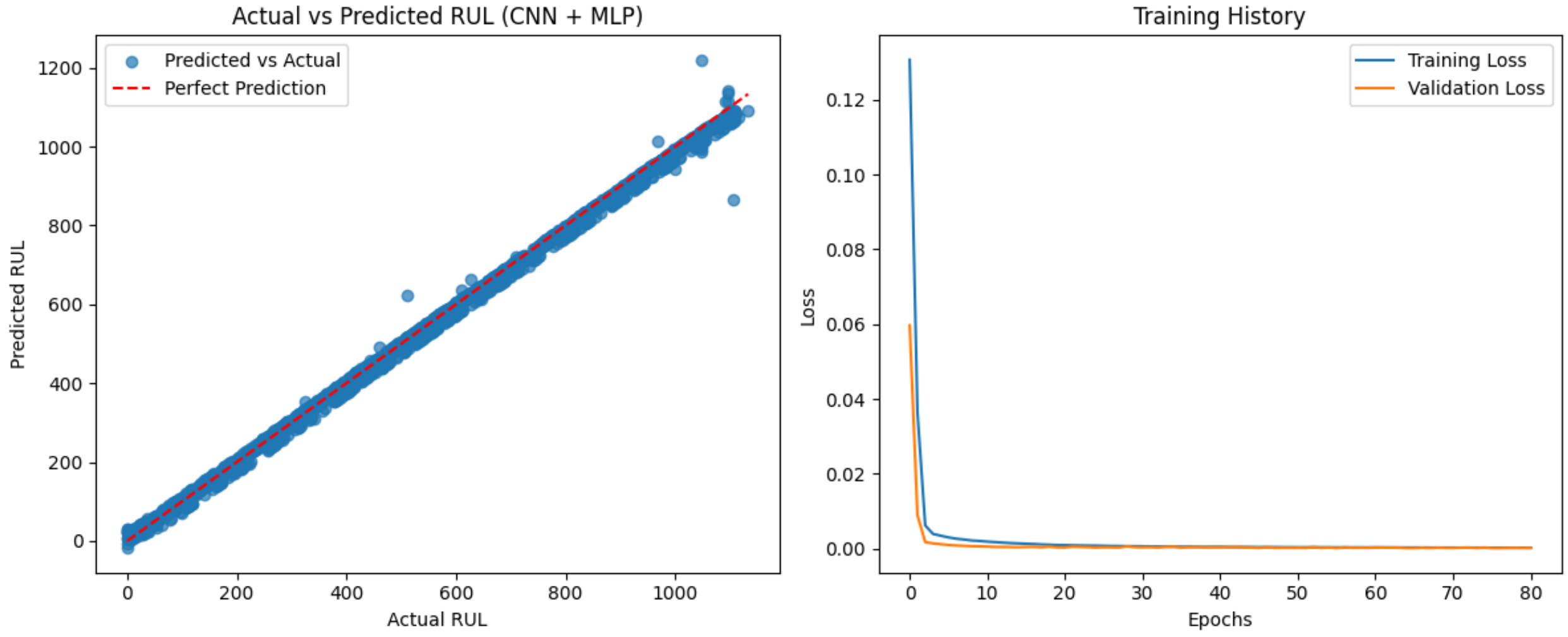

Figure 10: Model actual vs predicted plot and training history plot

Additionally, the model achieved an MAE of 8.0108, which, while not outperforming some other models, still highlights the potential of hybrid architectures in predictive tasks such as battery performance estimation. Fig 10 (right) presents the training and validation loss curves, showing a rapid convergence during the initial epochs, followed by a plateau at minimal loss values. This indicates effective training without significant overfitting, as evidenced by the close alignment of the training and validation loss curves. Overall, these results emphasize the promise of hybrid CNN-MLP models in predictive maintenance and performance forecasting applications. The detailed architecture of the model is provided in Figure 11.

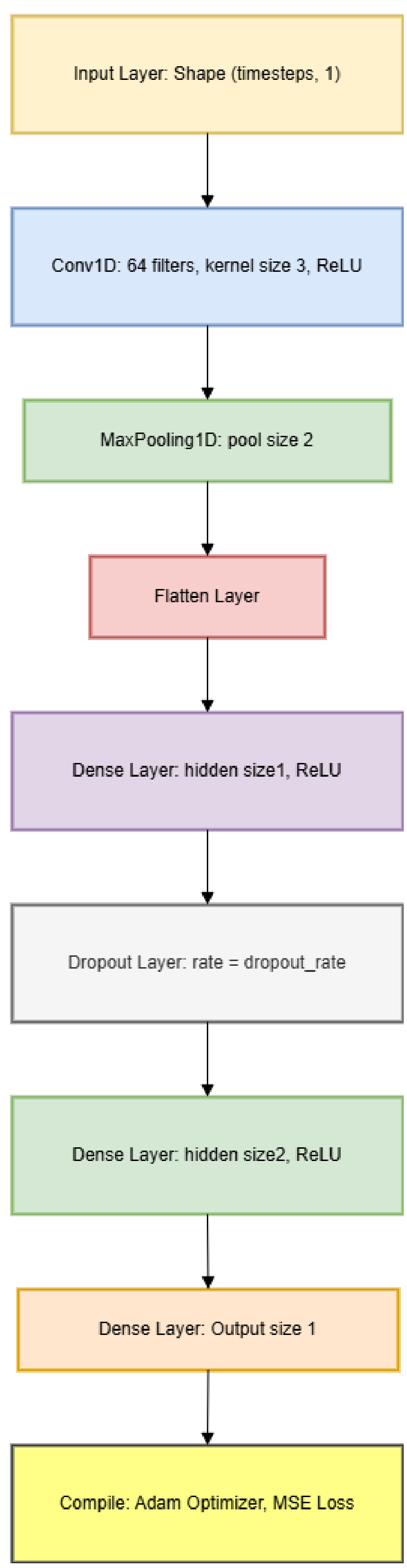

Figure 11: Hybrid model architecture

### *CNN-BiLSTM*

The CNN-BiLSTM hybrid model demonstrated strong predictive performance, achieving an MAE of 6.0099, an RMSE of 19.6879, and an $R^2$ score of 0.9963. As shown in Fig 12 (left), the actual vs. predicted RUL plot illustrates that most data points (blue dots) are closely aligned with the diagonal red line, representing perfect predictions. This alignment highlights the model's ability to accurately predict RUL across the dataset, although minor deviations are observed at certain points, particularly at extreme RUL values.

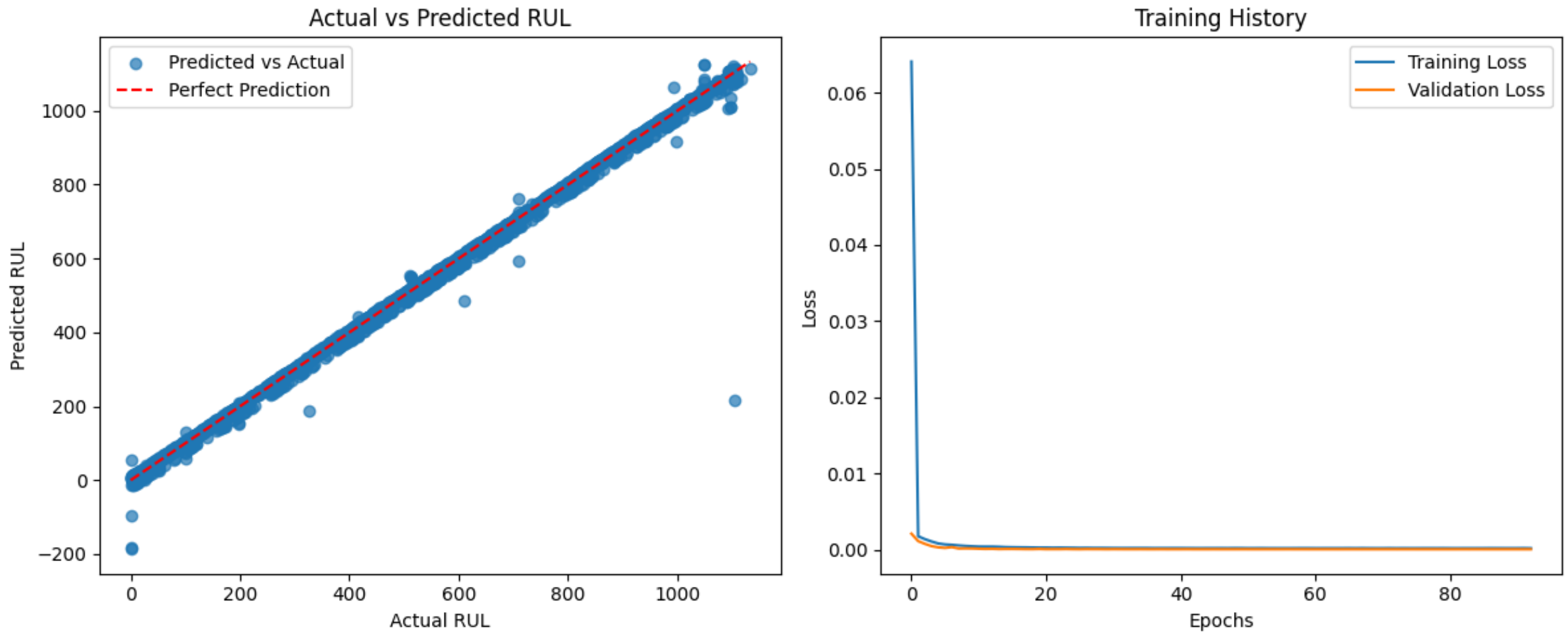

Figure 12: Model actual vs predicted plot and training history plot for CNN-BiLSTM

Fig 12 (right) presents the training and validation loss curves throughout training. The rapid convergence of the loss values within the first few epochs, followed by a steady plateau, indicates effective training and minimal overfitting. The close alignment of the training and validation loss further confirms the model's generalization capability. Overall, the CNN-BiLSTM model effectively combines feature extraction and sequential learning to achieve high accuracy in RUL prediction, showcasing its potential for predictive maintenance applications.

### *BiGRU*

The BiGRU model demonstrated excellent predictive accuracy, achieving an MAE of 6.0511, an RMSE of 11.1284, and an $R^2$ score of 0.9988. As shown in Figure 13 (left), the actual vs. predicted RUL plot illustrates that the majority of data points closely align with the diagonal red line representing perfect predictions. This alignment indicates that the model is highly effective at capturing the underlying patterns in the data and delivering accurate RUL predictions.

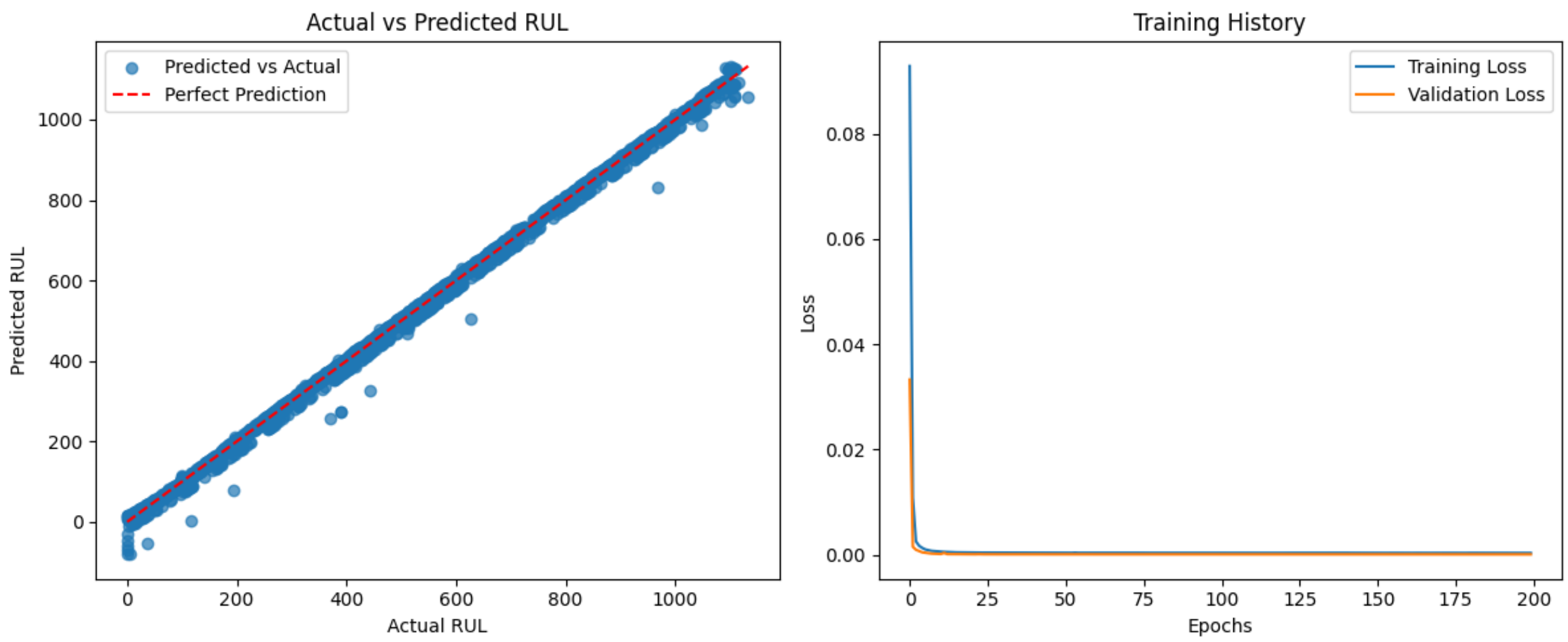

Figure 13: Model actual vs predicted plot and training history plot for BiGRU

Figure 13 (right) depicts the training and validation loss curves, where both losses converge rapidly during the initial epochs and plateau at low values, highlighting the efficient training process. The minimal gap between the training and validation loss further demonstrates the model's strong generalization capability. Overall, the BiGRU model exhibits robust performance in RUL prediction tasks, making it well-suited for predictive maintenance applications requiring high accuracy and reliability.

***TLE***

The two-level stacking and bagging ensemble model demonstrated outstanding performance, achieving an $R^2$ score of 0.9999, an RMSE of 2.8350, and an MAE of 1.4858. As shown in the actual vs. predicted plot in Fig 14, the blue data points align almost perfectly with the diagonal red line, which represents ideal predictions. This near-perfect alignment reflects the model's exceptional accuracy and minimal prediction error across all RUL values.

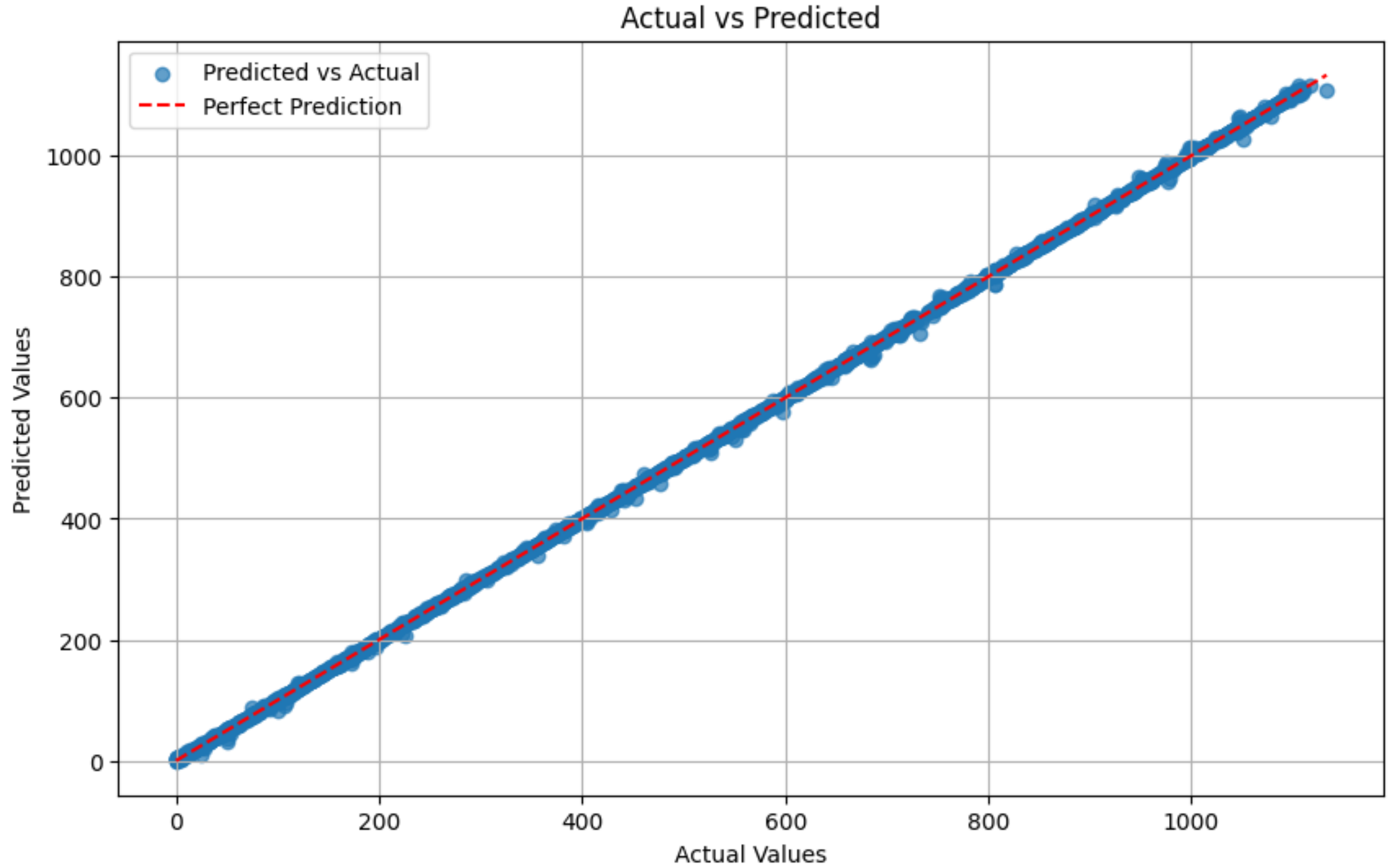

Figure 14: Actual vs predicted plot for TLE model

The superior performance of this ensemble model underscores the effectiveness of combining multiple base learners through stacking and bagging to capture complex patterns in the data while minimizing overfitting. With its ability to generalize well and achieve remarkably low error metrics, the two-level ensemble approach is highly reliable for predicting RUL in critical predictive maintenance applications.

***Classification models:***

***i. XGboost***

The XGBoost model obtained a testing accuracy testing accuracy: 0.9973. The model confusion matrix below shows the model performance on predicting various classes in which class 'high' is predicted correctly in 974 instances of the total testing set and out of 977 true instances. Similarl y, the class 'low' is predicted correctly for 1021 instances out of a total of 1025 true instances, an d finally class 'mid' is predicted correctly for 1005 instances out of a total of 1011 instances as s een in Fig 15.

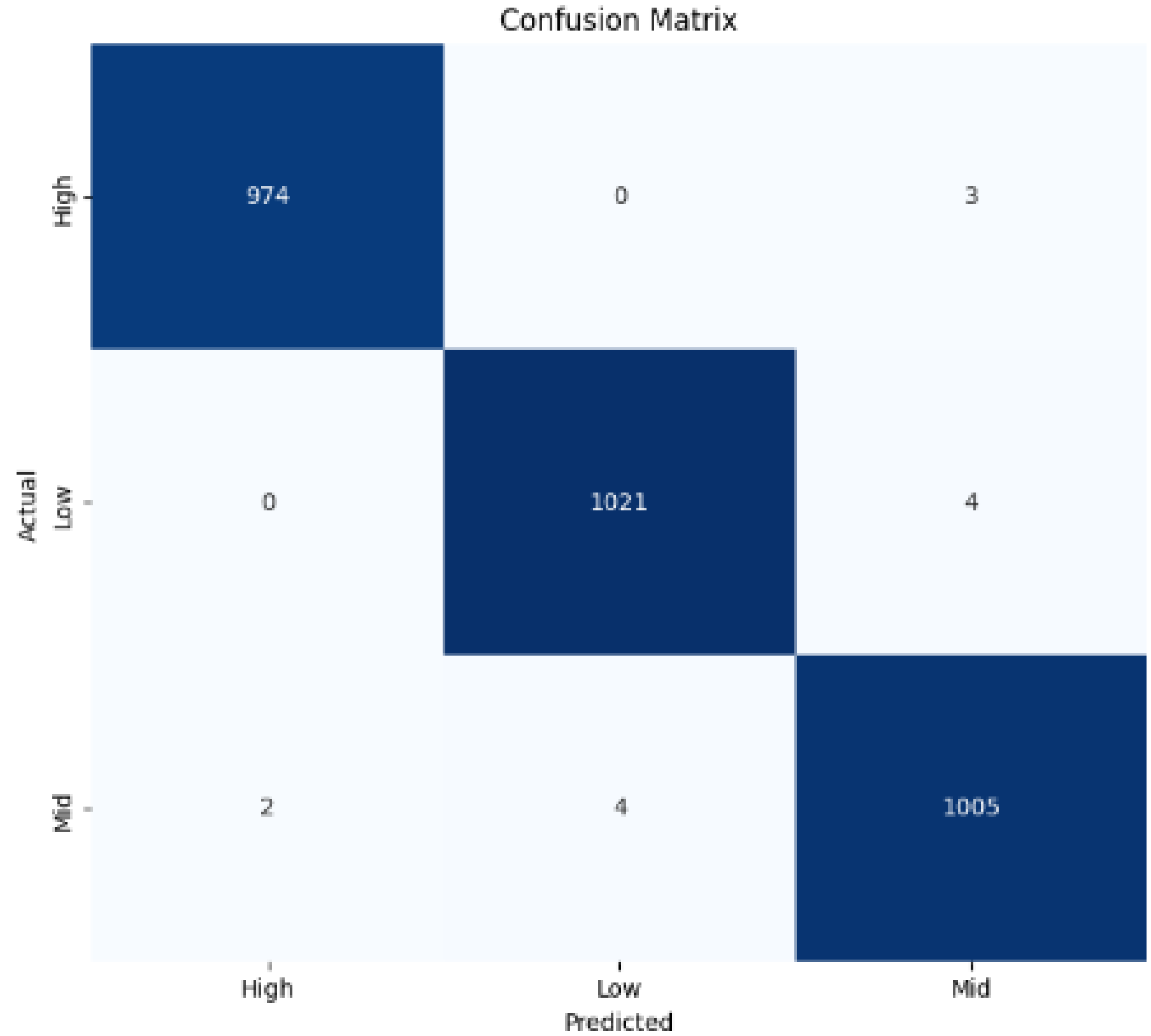

Figure 15: Confusion matrix plot for XGboost classifier

The confusion matrix-based classification and performance check shows how well the model fits to overall testing data.

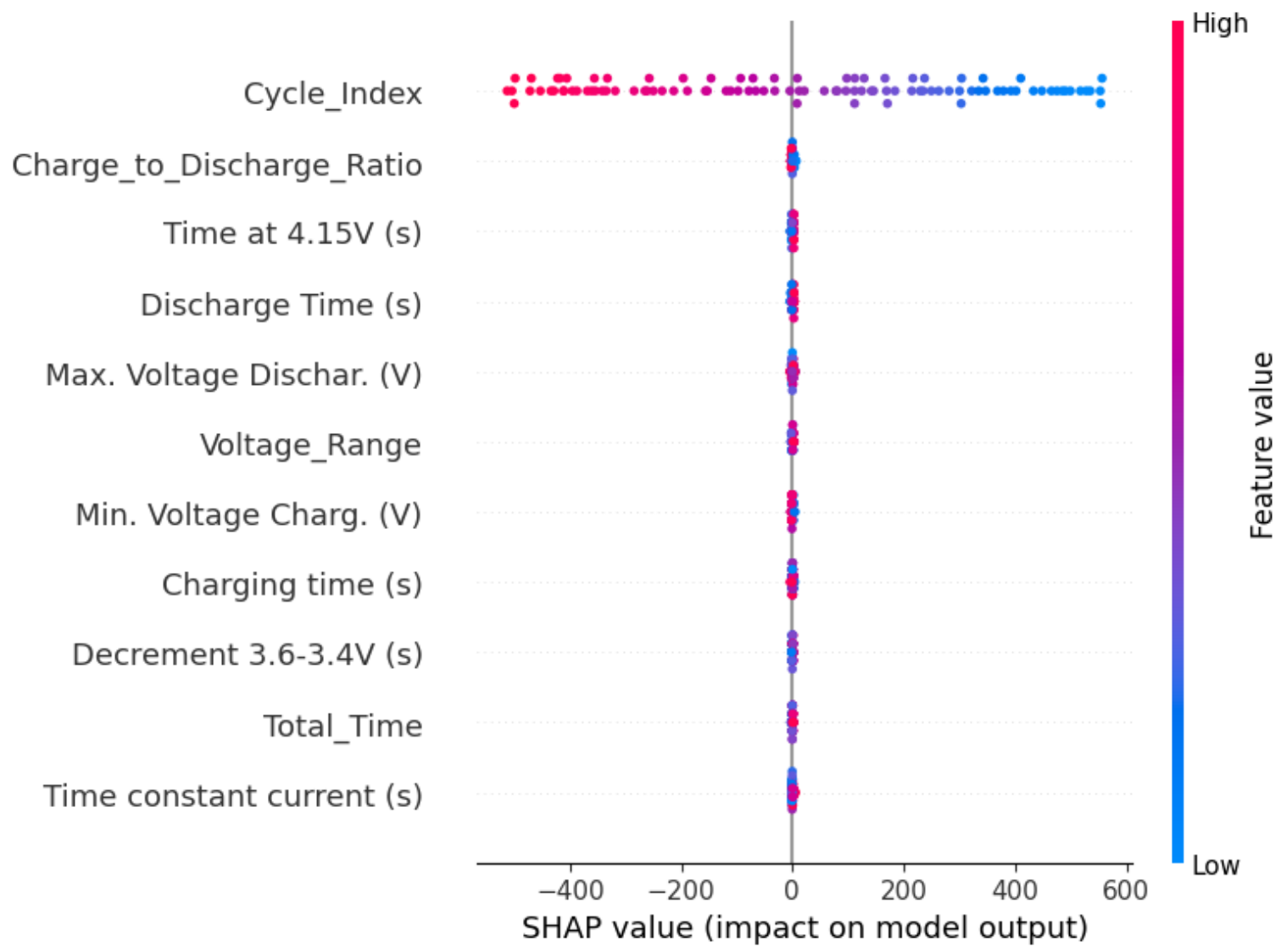

Figure 16: SHAP summary plot metrics result

Fig 16 presents a SHAP summary plot illustrating the feature contributions to the model's predictions, ranked by their average impact. The “Cycle_Index” emerges as the most influential feature, with a wide spread of SHAP values indicating its significant role in the predictions. “Charge_to_Discharge_Ratio” also shows notable importance, with higher values positively correlating with the output. Features like Time at 4.15V and Discharge Time exhibit moderate influence, while time-based metrics (e.g., “Charging time”, and “Total Time”) have comparatively smaller impacts. The color-coded feature values reveal non-linear effects and correlations, offering insights into feature relationships and their relative significance for model performance.

This bar plot, Fig 17, illustrates the average SHAP value magnitude for each feature, representing their relative importance in the model's predictions. The Cycle_Index dominates as the most significant feature, with a substantially higher mean SHAP value compared to all others, indicating its critical role in determining model outputs. Other features, such as Charge_to_Discharge_Ratio and Time at 4.15V, have minimal contributions, suggesting that the model is heavily reliant on Cycle_Index. This underscores the importance of focusing on Cycle_Index for understanding and improving the model's performance.

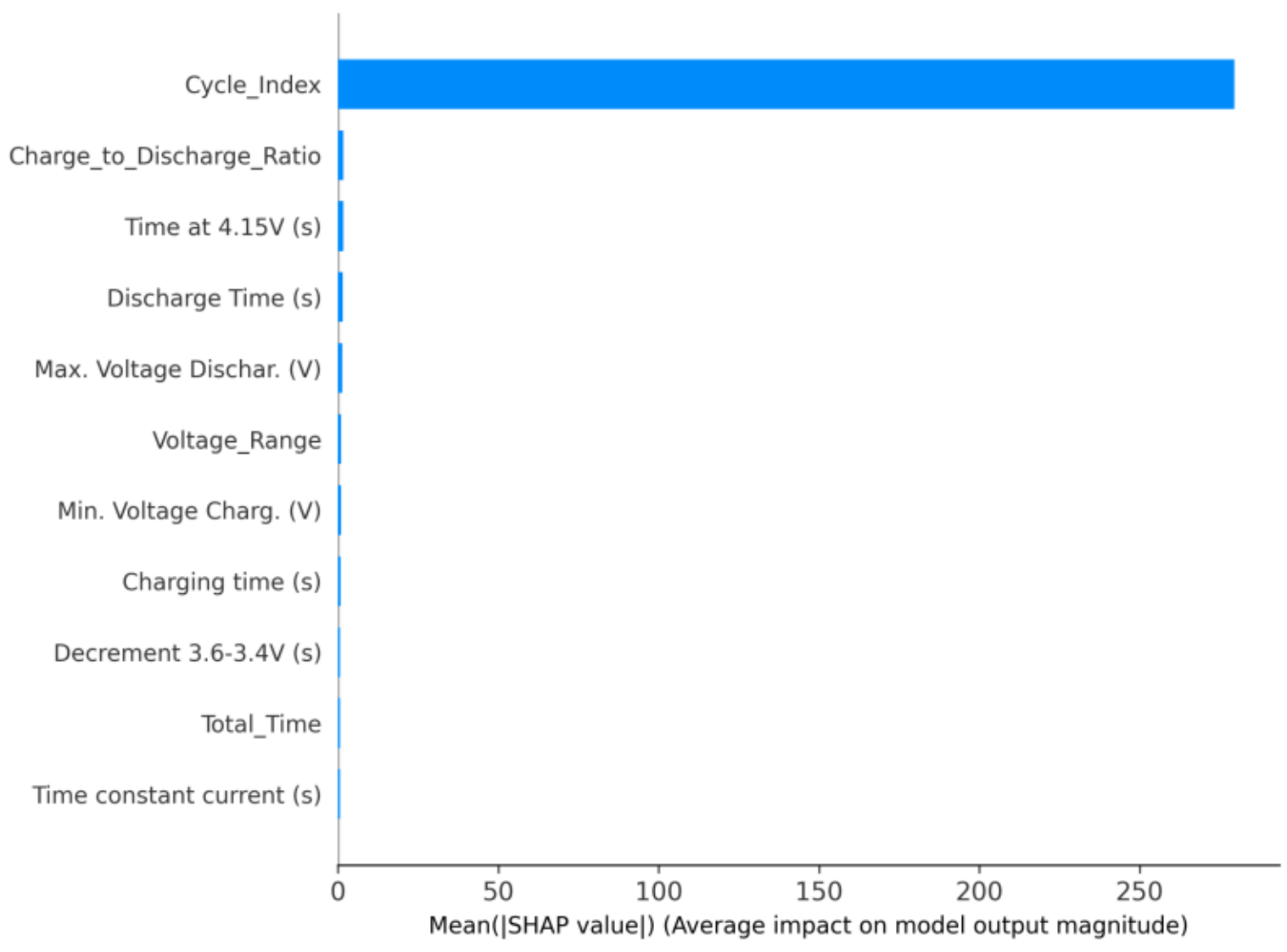

Figure 17: SHAP feature importance plot

***MLP model results:***

The MLP model obtained a testing accuracy testing accuracy: 0.0.9877 and a training accuracy o f 0.9903. The model confusion matrix in Fig 18 below shows the model performance on predicti ng various classes in which class 'high' is predicted correctly in 965 instances of total testing set and out of 977 true instances. Similarly, the class 'low' is predicted correctly for 1013 instances out of a total of 1025 true instances, and finally class 'mid' is predicted correctly for 998 instance s out of a total of 1011 instances. The confusion matrix-based classification and performance che ck shows how well the model fits to overall testing data.

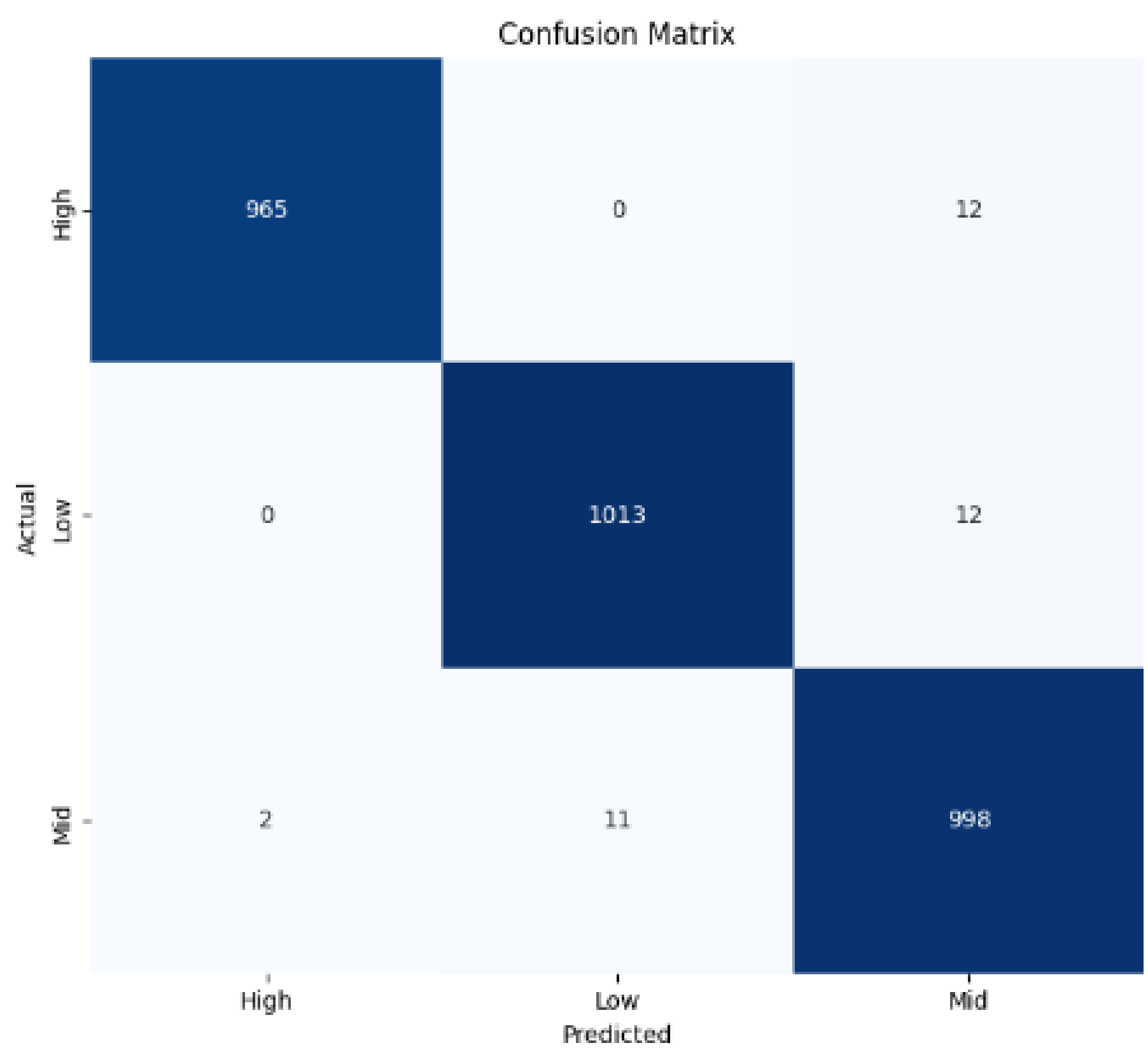

Figure 18: Confusion matrix plot for MLP model

***Model training and loss history plots:***

Fig 19 shows the history plot. The model fits well with both the training and validation data, showing strong learning performance over the epochs. The training and validation accuracy curves align closely and stabilize at high values, indicating that the model generalizes effectively to unseen data. Similarly, the training and validation loss curves decrease rapidly during the initial epochs and plateau at low values, confirming that the model learns efficiently without overfitting. The model trains for 38 epochs before early stopping triggers, ensuring that training halts once the validation performance stops improving. This consistent performance across training and validation highlights the model's robustness and its ability to extract meaningful patterns from the data.

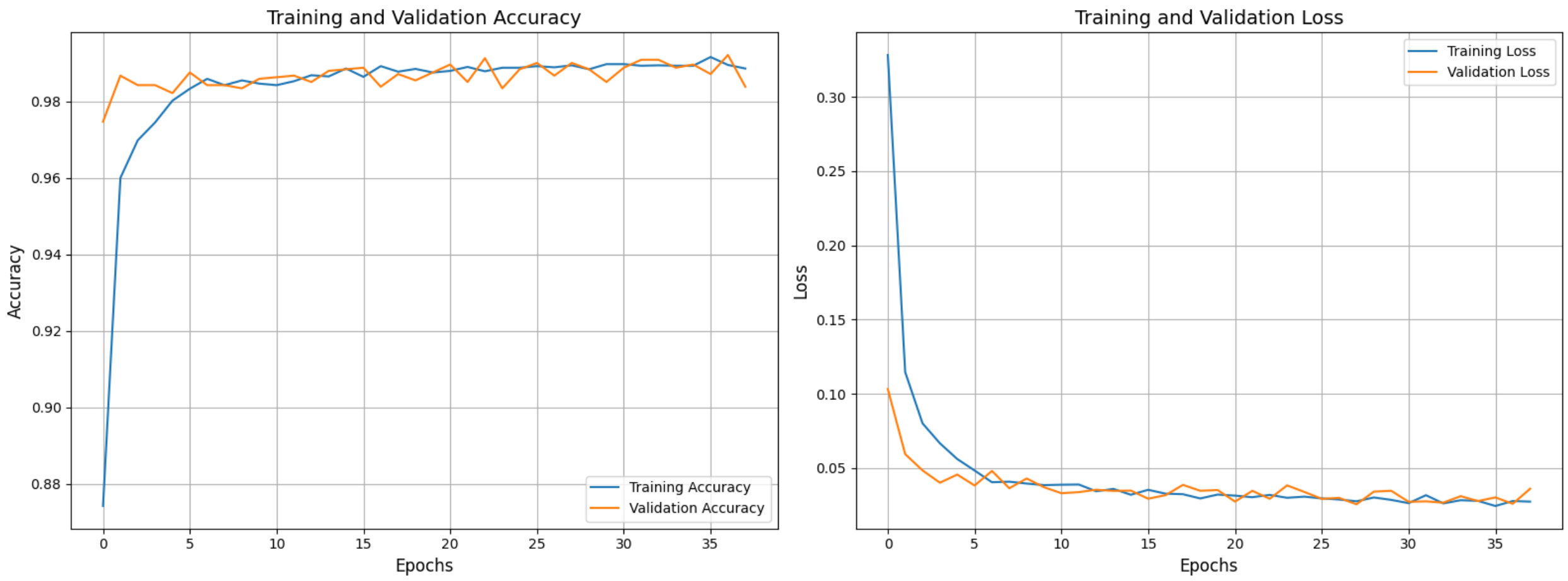

Figure 19: History plot for MLP model

**Models hyperparameters:**

Table 3: Best hyperparameters utilized for development

| Model | Hyperparameter | Value |
| --- | --- | --- |
| **XGBoost Regressor** | Colsample by Tree | 0.9984 |
| | Learning Rate | 0.0157 |
| | Max Depth | 11 |
| | Number of Estimators | 711 |
| | Subsample | 0.6035 |
| **CatBoost Regressor** | Border Count | 255 |
| | Depth | 3 |
| | Iterations | 1000 |
| | L2 Leaf Regularization | 6.8461 |
| | Learning Rate | 0.3 |
| **MLP + CNN Model** | Hidden Size 1 | 124 |

| | Hidden Size 2 | 128 |
|---|---|---|
| | Learning Rate | 0.0002 |
| | Number of Epochs | 200 |
| | Batch Size | 122 |
| | Dropout Rate | 0.3279 |
| **CNN + BiLSTM** | Conv1D layer | 4 |
| | MaxPooling | 2 |
| | BiLSTM | 104 |
| | Dense | 40 |
| | Dropout | 0.2 |
| | Output | 1 |
| | Optimizer | Adam |
| | Learning rate | 0.0005 |
| | Loss function | MSE |
| | Batch Size | 64 |
| | Epochs | 150 |
| **BiGRU** | BiGRU layer 1 | 128 |
| | BiGRU layer 2 | 64 |
| | Dropout | 0.4 |
| | Dense | 32 |
| | Output | 1 |
| | Optimizer | Adam |
| | Learning rate | 0.0001 |
| | Loss function | MSE |
| | Batch Size | 64 |
| | Epochs | 200 |
| **XGBoost Classifier** | Colsample by Tree | 0.5 |
| | Gamma | 0.0 |
| | Learning Rate | 0.3 |
| | Max Depth | 9 |
| | Min Child Weight | 1 |
| | Number of Estimators | 500 |
| | Reg Alpha | 0.0 |
| | Reg Lambda | 0.063 |
| | Subsample | 1.0 |
| **MLP Classifier** | Epochs | 200 |
| | Batch Size | 32 |
| | Early Stopping Patience | 10 |
| | Validation Split | 0.2 |
| | Hidden Layer 1 | 128 units (ReLU) |
| | Hidden Layer 2 | 64 units (ReLU) |
| | Output Layer | 3 units (Softmax) |
| | Dropout | 0.3 |

Table 4: Best hyperparameters for TLE

| **Model** | **Hyperparameter** | **Value** |
|---|---|---|
| **Decision Tree** | Min Samples Split | 2 |
| | Min Sample Leaf | 2 |
| | Max Depth | None |
| **Extra Trees** | Number of Estimators | 200 |
| | Min Samples Split | 2 |
| | Min Samples Leaf | 1 |
| | Max Depth | None |
| **Random Forest** | Number of Estimators | 200 |
| | Min Samples Split | 2 |
| | Min Samples Leaf | 1 |
| | Max Depth | None |
| **XGBoost** | Subsample | 1.0 |
| | Number of Estimators | 500 |
| | Max Depth | 10 |
| | Learning Rate | 0.1 |
| | Colsample by Tree | 1.0 |
| **LightGBM** | Subsample | 1.0 |
| | Number of leaves | 100 |
| | Number of Estimators | 500 |
| | Max Depth | 20 |
| | Learning Rate | 0.1 |
| **CatBoost** | Learning Rate | 0.2 |
| | Iterations | 500 |
| | Depth | 4 |
| **Bagging Regressor** | Base Estimator | Linear Regression |
| | Number of Estimators | 10 |

**Models Performance comparison:**

**Table 5:** Model performance comparisons

| Model | Metric | Value |
|---|---|---|
| **XGBoost Regressor** | $R^2$ Score | 0.9994 |
| | RMSE | 8.0441 |
| | MAE | 4.4955 |
| **CatBoost Regressor** | $R^2$ Score | 0.9980 |
| | RMSE | 14.352 |
| | MAE | 10.607 |
| **MLP + CNN Model** | $R^2$ Score | 0.9985 |
| | RMSE | 12.632 |
| | MAE | 8.0108 |
| **CNN+BiLSTM** | $R^2$ Score | 0.9963 |
| | RMSE | 19.687 |
| | MAE | 6.0099 |

| **BiGRU** | $R^2$ Score | 0.9980 |
|---|---|---|
| | RMSE | 14.222 |
| | MAE | 9.7447 |
| **TLE Regressor** | $R^2$ Score | **0.9999** |
| | RMSE | **2.8350** |
| | MAE | **1.4858** |
| **XGBoost Classifier** | Training Accuracy | **0.9998** |
| | Testing Accuracy | **0.9957** |
| **MLP Classifier** | Training Accuracy | 0.9903 |
| | Testing Accuracy | 0.9877 |

Table 6: Classification 10-fold-cross-validation

| Fold | XGBoost | MLP |
|---|---|---|
| 1 | 0.9973 | 0.9960 |
| 2 | 0.9960 | 0.9907 |
| 3 | 0.9973 | 0.9854 |
| 4 | 0.9960 | 0.9914 |
| 5 | 0.9973 | 0.9934 |
| 6 | 0.9973 | 0.9927 |
| 7 | 0.9947 | 0.9861 |
| 8 | 0.9947 | 0.9914 |
| 9 | 0.9987 | 0.9914 |
| 10 | 0.9960 | 0.9920 |
| **Mean Accuracy** | **0.9969** | **0.9969** |
| **Standard Deviation** | **0.0016** | **0.0016** |

Table 7: 5-fold Stratified Cross-validation

| Fold | Metric | XGBoost | CatBoost | CNN + MLP | CNN+BiLSTM | BiGRU | TLE |
|---|---|---|---|---|---|---|---|
| **1** | **MAE** | 4.1747 | 10.3668 | 4.9180 | 6.6901 | 6.1138 | 1.6427 |
| | **RMSE** | 6.7875 | 14.2640 | 8.4860 | 12.5039 | 10.2122 | 3.0410 |
| | **$R^2$** | 0.9996 | 0.9980 | 0.9993 | 0.9985 | 0.9990 | 0.9999 |
| **2** | **MAE** | 4.3791 | 10.7548 | 9.4825 | 5.7928 | 7.1393 | 1.6129 |
| | **RMSE** | 8.1498 | 14.3145 | 15.9910 | 12.6643 | 11.2259 | 2.9747 |
| | **$R^2$** | 0.9994 | 0.9981 | 0.9976 | 0.9985 | 0.9988 | 0.9999 |
| **3** | **MAE** | 4.2851 | 10.4645 | 7.8121 | 5.9341 | 5.6390 | 1.4921 |
| | **RMSE** | 7.0519 | 12.9722 | 12.6418 | 11.2106 | 9.9236 | 2.6842 |
| | **$R^2$** | 0.9995 | 0.9984 | 0.9985 | 0.9988 | 0.9990 | 0.9999 |
| **4** | **MAE** | 4.2261 | 10.6111 | 4.8276 | 5.7538 | 5.5194 | 1.5302 |
| | **RMSE** | 7.2339 | 14.5824 | 7.7936 | 8.7421 | 8.9666 | 2.8849 |
| | **$R^2$** | 0.9995 | 0.9979 | 0.9994 | 0.9993 | 0.9992 | 0.9999 |
| **5** | **MAE** | 4.0753 | 10.1435 | 5.1656 | 4.9148 | 6.0037 | 1.5777 |
| | **RMSE** | 5.8180 | 12.5937 | 10.1854 | 8.1566 | 9.7256 | 2.7959 |
| | **$R^2$** | 0.9997 | 0.9985 | 0.9990 | 0.9994 | 0.9991 | 0.9999 |
| **Mean** | **MAE** | 4.2281 | 10.4682 | 6.4412 | 5.8171 | 6.0830 | **1.5711** |
| | **RMSE** | 7.0082 | 13.7453 | 11.0196 | 10.6555 | 10.0108 | **2.8761** |
| | **$R^2$** | 0.9995 | 0.9982 | 0.9987 | 0.9989 | 0.9990 | **0.9999** |
| **SD** | **MAE** | 0.1022 | 0.2089 | 1.8804 | 0.5647 | 0.5723 | **0.0544** |
| | **RMSE** | 0.7512 | 0.8022 | 2.9956 | 1.8796 | 0.7343 | **0.1267** |
| | **$R^2$** | 0.0001 | 0.0002 | 0.0007 | 0.0004 | 0.0001 | **0.0000** |

From Table 5, it is clear that the best performance is given by the proposed TLE model for predicting RUL with RMSE of 2.8350, MAE 1.4858, and $R^2$ 0.999, further validated with cross-validation as shown in Tables 6 and 7. Similarly, XGBoost classifier obtained the best performance for classifying the RUL for the battery. The models are optimized using Bayesian optimization and the best hyperparameters found are shown in Tables 3 and 4. These performance measures show how well various machine learning and neural network algorithms fit the data in predicting the RUL for batteries.

The TLE demonstrates superior performance, particularly for tabulated data, due to its two-level stacking approach, which effectively combines multiple base models to capture complex relationships within the dataset. This ensemble technique enhances predictive accuracy by leveraging diverse learning patterns from individual models at the first level and integrating them optimally at the second level. As a result, the TLE model achieves the highest $R^2$ score and the lowest RMSE and MAE, proving its robustness and efficiency in handling structured data for accurate battery RUL prediction.

The GUI shown in Figure 22(a) offers a great advantage at various levels, with the help of the testing data sett we successfully insert data to the application and thus, it helped to visualize the model predictions on unseen data.

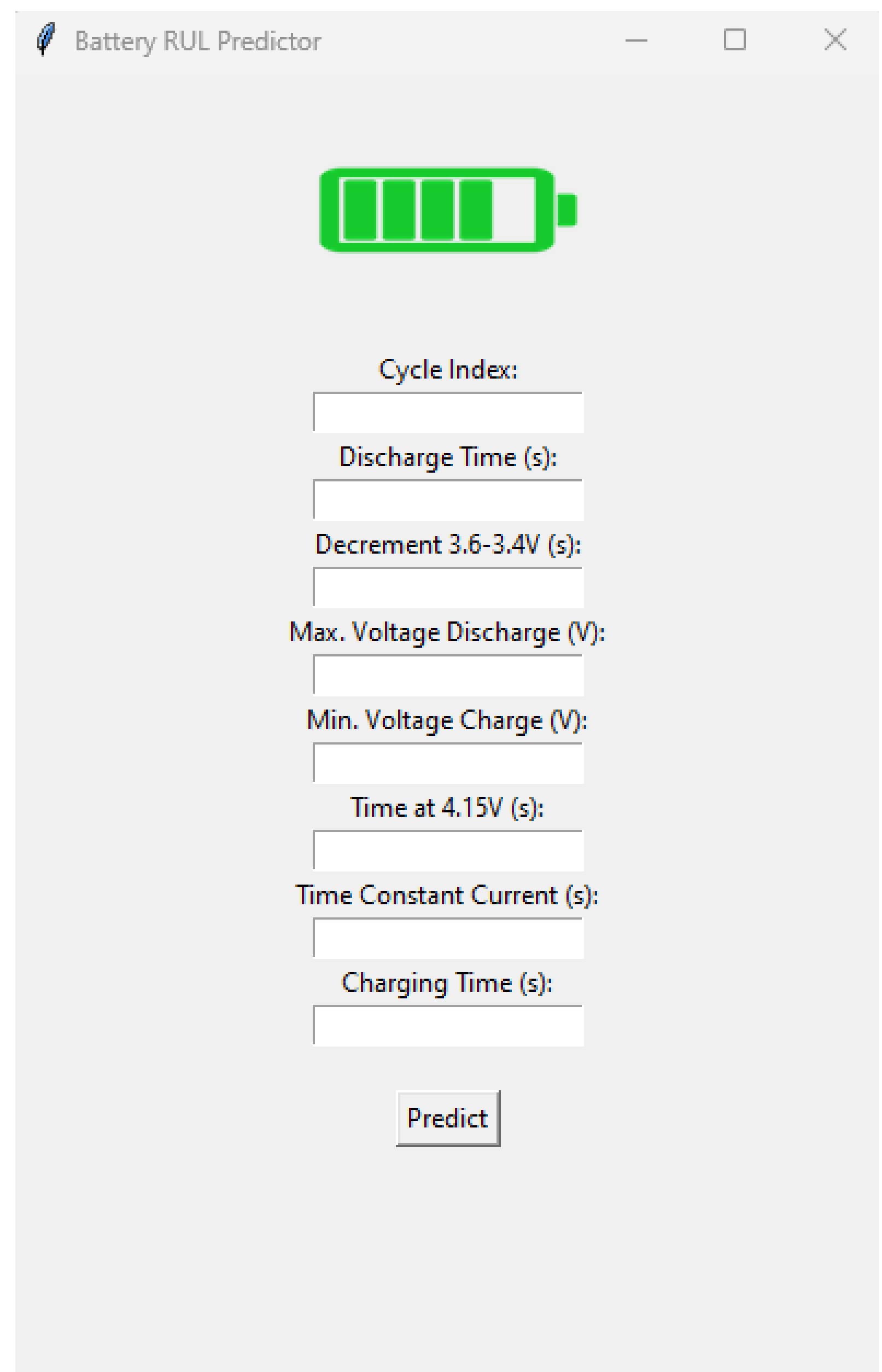

Fig 22(a). GUI for application

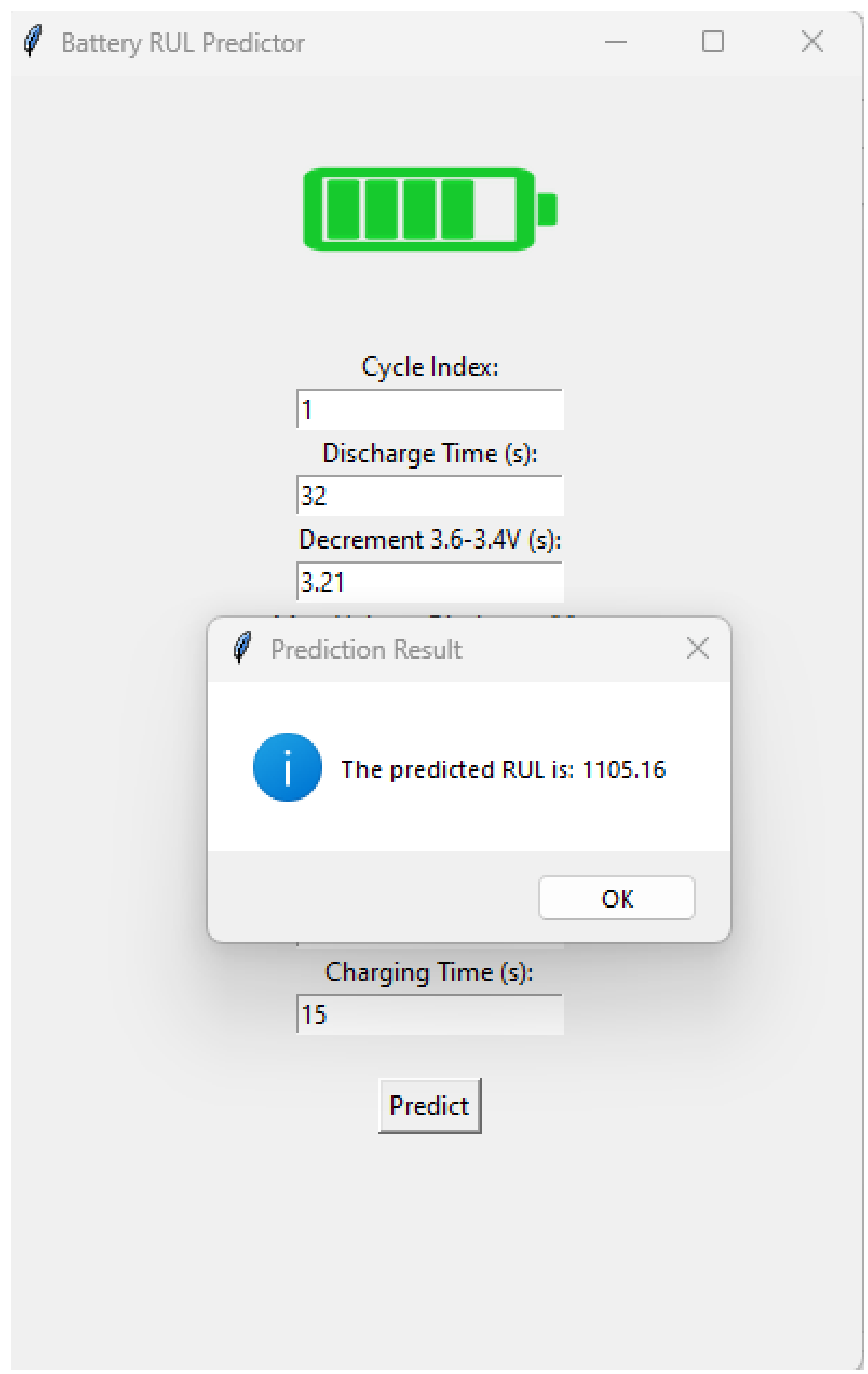

**Fig 22 (b).** GUI-based predictions made

Figure 22 (b). shows the battery-related parameters input GUI system that is integrated to the used to make predictions on battery RUL similarly, intergarting a classifier model can also be used in the GUI and it can be upgraded to classify the battery life. As the system does not natively support the model, we integrated functionality into the backend to enable it to input testing data and actuate the relay using the 'micromlgen' library. This implementation demonstrates how the model operates within the system to trigger charging when needed. Through this simulation-based approach, we showcase how machine learning-driven automated charging systems predicts RUL levels and initiate recharging processes autonomously, which is the primary objective of the system's development. The GUI shown in Figure 22(c) supports the classification approach for battery RUL predictions.

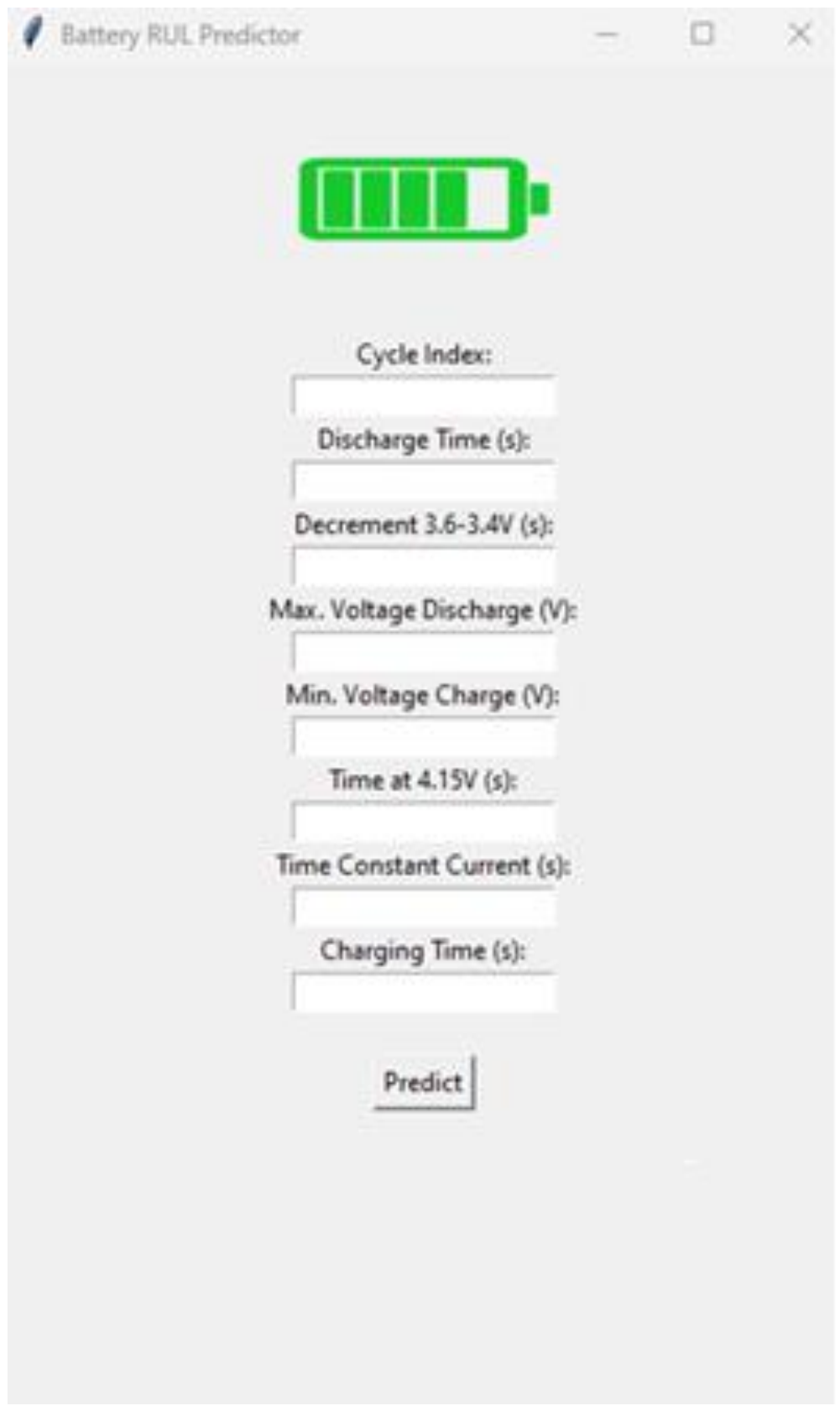

**Figure 22(c):** GUI for classification of RUL

### 3.2 Comparison with previous works:

Table 8: Comparison of results with previous works

| Reference | Method | Result |
| --- | --- | --- |
| [1] | Explainable, real-time machine learning for battery | Insights into challenges with in situ computations and data gathering. |
| | production and optimization. | |
| [2] | Machine learning techniques like SVM, FL, KNN, and GA for predicting SOC and SOH. | Highlights difficulties in battery management systems; emphasizes the use of various algorithms. |
| [3] | Neural network-based prediction method for SOC and SOH estimates. | The best performance noted with high accuracy and low RMSE for electric vehicle applications. |
| [4] | Ensemble random forest model for RL prediction, incorporating data gathering and preprocessing. | Higher prediction accuracy with $R^2$ and RMSE metrics. |
| [5] | Comparative analysis of LSTM and Bi-LSTM for forecasting LIB performance. | Evaluates MSE, MAE, RMSE, and R-squared metrics; aims to advance electric vehicle technology. |
| [6] | Decision trees, random forests, and linear regression for SoC prediction using Panasonic Li-Ion data. | Random forest regressor outperformed with a correlation coefficient of 0.9988. |
| [7] | Advanced ML techniques for SOC forecasting with regression under dynamic loads. | Showcases superior predictive capability of advanced ML models over standard approaches. |
| [8] | Feature extraction with multiple linear regression to predict full battery charge curve. | Achieved less than 2% prediction error using only 10% of the charge curve data. |
| [9] | Probabilistic machine learning models for battery | Discusses uncertainty quantification and prospects for future research. |

| | health diagnostics and prognostics. | |
|---|---|---|
| [10] | LSTM, DT, KNN, NB, SVM used for real-time battery data prediction. | Naïve Bayes yielded the best results with 88% accuracy for predicting remaining battery capacity. |
| [11] | MLP model evaluated for extrapolation accuracy in predicting battery voltage. | Best extrapolation accuracy noted with a value of 92.7 mV. |
| [12] | Physics-based and ML modeling techniques for SoC predictions at high C-rates. | Pseudo-2D electrochemical model estimated SoC within ~2% RMSE; feed-forward neural network had RMSE <1%. |
| **Our Work** | Developed ML models for predicting ((TLE, MLP+CNN) and classifying (XGBoost, MLP) battery RUL; SHAP interpretability analysis; | Achieved 99% accuracy (XGBoost, 98% (MLP); TLE showed RMSE of 2.8350, MAE 1.4858 and $R^2$ 0.9999 for regression and insights into critical factors like cycle index and charging parameters for sustainable battery management and interpretability analysis performed using SHAP explainable AI, along with the GUI application developed for predicting the RUL for batteries. |

The related works cover a range of machine-learning methods used to predict various battery conditions and performance. While tackling issues in battery management systems, they examine techniques like support vector machines (SVM) [2], neural networks [5], and ensemble models, emphasizing how well they predict the state of charge (SOC), state of health (SOH) [7], and remaining life (RL). Important conclusions include the usage of sophisticated methods like LSTM and random forests for enhanced prediction skills, as well as high accuracy and low root mean square error (RMSE) metrics, especially in electric car applications.

In contrast to these works, we focus on creating machine learning models to predict and classify battery RU. In addition to a graphical user interface (GUI) for useful RUL predictions, we also performed interpretability analysis using SHAP and reached up to 99% accuracy with our models. The use of explainable AI offers a solid concept for the development of an RUL prediction related most affecting factors [29, 30]. Finally, this research work supports a vital aspect of sustainable battery management utilizing ML as well as making useful decisions with the help of a GUI application developed for predicting the battery RUL.

### *3.3 Limitations and future scopes:*

The work has several limitations, like the need for research for more reliable and valid data from rigorous testing. Additionally, the app system currently lacks real-world implementation capabilities. Overall, the system shows an example configuration of how ML can be utilized to develop RUL automated charging systems. The utilized various hardware components and their usage, as well as operating voltage, are given in Table 9.

In the future, we aim to deploy the model on a Raspberry Pi system and enhance it with a real-life data extraction-based prototype for developing the system that is directly fed to the data acquisition from the battery along with a recharging system for power management. The IoT-based automation can be developed by creating its database and server for connecting to the hardware system and managing remotely for power-related tasks. The system can be developed with powerful microcontrollers and actuators that can have real-time implementation in charging systems and power management. The integration technique to IoT can be integrated in real-time and deployed in real-time systems automating the battery charging process. Overall, the system can be a great approach for energy management and saving in the future.

### *3.4 Future applications*

The developed Python model can have the following applications in modern society:

**a) Potential Applications in real-time appliances**

For power management and faulty system analysis, we can create many systems that can allow us to manage batteries and overlook the outcomes of the RUL, automate them to be intelligent enough for self-fault handling and managing the issues related to power management. This allows systems to not only rely on software but also to make physical connections and large power-based damage protections.

#### b) Optimization of Vehicle systems:

The created model can be used to improve the electric vehicle's performance. Manufacturers can anticipate projected power management by automating the recharging systems in power stations which makes more intelligent and automated charging systems useful for EV systems [13-25]. This makes it possible to improve design iteratively and improve overall performance and battery efficiency.

#### c) Energy saving mechanism

The system being able to adapt the change of battery RUL use of AI based automation can help to allow power management fault analysis and also could handle various hazard related to power consumption and over use.

#### d) Decision Support for Manufacturers:

Manufacturers can utilize the model for decision support during the development phase. By inputting various advanced microcontrollers and semiconductors, they can assess the potential charging outcomes, facilitating informed decisions on component selection, vehicle dimensions, and other critical aspects of fuel cell vehicle design.

#### e) Evaluating Real-world Scenarios:

The resilience of the model enables the assessment of real-world scenarios in addition to real-time data. Manufacturers and researchers can acquire estimates that inform judgments about the performance of fuel cell vehicles in real-world scenarios by entering existing or projected vehicle characteristics [15].

## 4. Conclusion

In light of this, power management systems based on AI offer hope for sustainable advancement in the field of rapidly advancing technology. The application of machine learning techniques facilitates the development of predictor procedures for determining the various parameters involved in the design and research of these charging systems for vehicles and other practical life applications. The excellent performance shown by the regression and classification models with 99% approximate accuracy shows the various models' potential for predicting battery-related life, and similarly, GUI shows the prediction capabilities with the help of an AI-model integrated approach. The SHAP interpretability analysis is a much more significant approach to identifying the factors affecting model predictions.

In conclusion, the developed model shows great promise for real-world applications in the design, optimization, and decision-making processes related to real-time power management and fault handling, in addition to its exceptional accuracy in predicting RUL data. This approach is particularly useful in applications where precise battery life estimation is crucial, such as in electric vehicles, energy storage systems, and remote sensing devices. Because of its adaptability, academics and different power sector and industry stakeholders can both benefit from it as a useful tool.

## Declaration of conflict of Interest

The Authors declare that there are no known conflicts of interest.

## Declaration of LLM use

The authors would like to declare that they utilized LLM tools for minimizing grammar errors and writing.

## Data and resources availability statement

The data is openly available on the kaggle platform. Also, dataset and codes can be accessed openly from the github repository as: Biplov01/codes-and-data-for-battery-XAI-deep-learning-system

**Declaration of conflict of interest:**

The authors declare that there are no known competing interests

**Authors contribution statement:**

- Biplov Paneru: Conceptualization, writing, methodology, model development, validation, analysis and validation, data curation
- Bipul Thapa: Proposed Solution, validation, model development. data curation, methodology, formal analysis, final draft preparation
- Durga Prasad Mainali: validation, data curation, methodology, formal analysis, final draft preparation
- Bishwash Paneru: Validation, Model Development, data curation, validation, methodology, software, final draft preparation